\documentclass{bmvc2k}
\usepackage{verbatim}

\title{ZeBRA: Precisely Destroying Neural Networks with
\underline{Ze}ro-Data \underline{B}ased \underline{R}epeated Bit Flip \underline{A}ttack}
\addauthor{Dahoon Park*}{pdh930105@dgist.ac.kr}{1}
\addauthor{Kon-Woo Kwon}{konwoo@hongik.ac.kr}{2}
\addauthor{Sunghoon Im}{sunghoonim@dgist.ac.kr}{1}
\addauthor{Jaeha Kung**}{jhkung@dgist.ac.kr}{1}

\addinstitution{
 Daegu Gyeongbuk Institute of Science and Technology (DGIST),\\
 Daegu, Korea
}
\addinstitution{
 Hongik University,\\
 Seoul, Korea
}
\runninghead{D. Park et al.}{Zero-Data Based Repeated Bit Flip Attack}

\usepackage{times}
\usepackage{epsfig}
\usepackage{graphicx}
\usepackage{amsmath}
\usepackage{amssymb}
\usepackage{comment}
\usepackage[ruled,vlined,linesnumbered]{algorithm2e}
\usepackage{algorithmic}
\usepackage{ctable}
\usepackage{multirow}

\newcommand{\Figref}[1]{Figure~\ref{#1}}
\newcommand{\Tabref}[1]{Table~\ref{#1}}

\newcommand{\tabref}[1]{Table ~\ref{#1}}
\newcommand{\eqnref}[1]{Eq.~(\ref{#1})}
\newcommand{\secref}[1]{Sec.~\ref{#1}}
\newcommand{\algref}[1]{Alg.~\ref{#1}}

\begin{document}

\maketitle

\begin{abstract}

   In this paper, we present Zero-data Based Repeated bit flip Attack (ZeBRA)
   that precisely destroys deep neural networks (DNNs) by synthesizing its own attack datasets.
   Many prior works on adversarial weight attack require not only
   the weight parameters, but also the training or test dataset in searching vulnerable
   bits to be attacked.
   We propose to synthesize the attack dataset, named distilled target data, 
   by utilizing the statistics of batch normalization layers in the victim DNN model.
   Equipped with the distilled target data, our ZeBRA algorithm can
   search vulnerable bits in the model without accessing training or test dataset.
   Thus, our approach makes the adversarial weight attack more fatal to the
   security of DNNs.
   Our experimental results show that 2.0$\times$ (CIFAR-10) and 1.6$\times$ (ImageNet) 
   less number of bit flips are required on average to destroy DNNs compared to the previous attack method. Our code is available at \url{https://github.com/pdh930105/ZeBRA}.

\end{abstract}

\section{Introduction}
\label{sec:intro}

\begin{figure}
    \centering
    \includegraphics[scale=0.4]{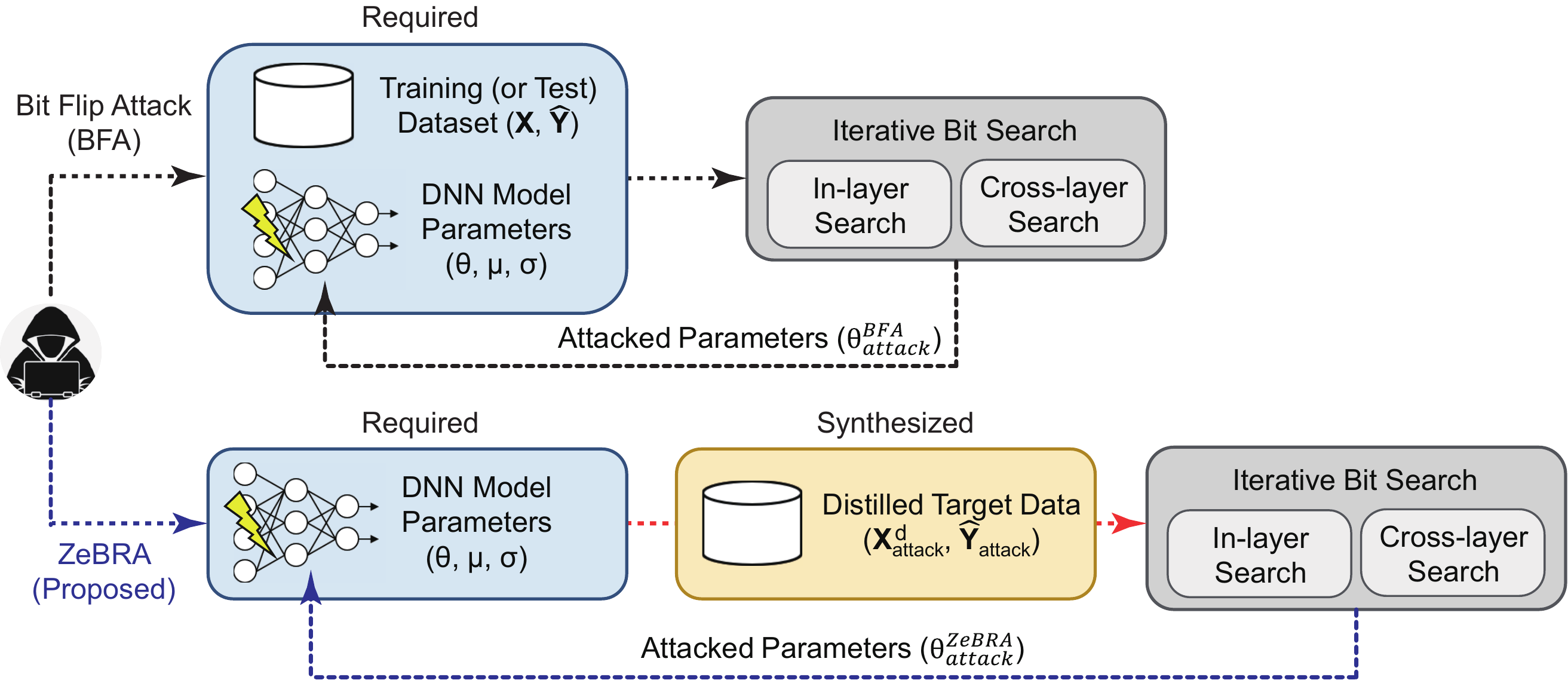}\vspace{-1mm}
    \caption{The overview of the proposed attack method (ZeBRA).
        The difference between the prior work~\cite{bfa} and ZeBRA is 
        the required access to the actual dataset for the attack.}
    \label{fig:intro}\vspace{-2mm}
\end{figure}

Recent advances in deep neural networks (DNNs) have led the
proliferation of DNN-assisted applications such as computer vision, machine translation, recommendation system, playing games, and robotics, 
to name a few~\cite{yolo,brown2020language,liu2020deep,youtube,mnih2013playing,huang2019learning}.
Moreover, as safety-critical applications are widely adopting deep learning,
i.e., medical imaging~\cite{nature2020kaissis}, self-driving cars~\cite{Grigorescu_2020}, and intelligent robots~\cite{melis2017deep},
the robustness of DNN models is getting extremely important.
For instance, an adversary can alter the behavior of the DNN model deployed in a self-driving car
to misclassify traffic signs~\cite{morgulis2019fooling}.
Thus, deep learning researchers need to carefully identify and understand
the unexpected blind spots of DNNs.
There are two different ways of attacking the DNN model:
i) adding imperceptible noise to input data ({\it adversarial examples}) and 
ii) moving decision boundaries by changing the weight parameters ({\it adversarial weight attack}).
Adversarial examples are trained/optimized to move away from the correct labels
for classification tasks~\cite{adv_ex_15}.
These examples have the attacking ability even after 
they are printed and photographed with a smartphone~\cite{kurakin2017adversarial}
or fed into other DNNs with different parameters and/or architectures~\cite{adv_ex_15}.
On the contrary, the adversarial weight attack changes the values of 
weight parameters by flipping bits of the DNN model~\cite{bfa, tbfa}.
The prior work, named bit flip attack (BFA), presents
an efficient way of finding vulnerable bits in the DNN model
via iterative bit search (\Figref{fig:intro}).

To perform the iterative search for finding bits to be flipped, 
the BFA requires DNN model parameters, i.e., weight ($\theta$) 
and batch normalization parameters ($\mu$, $\sigma$), and the training or test dataset.
An adversary may have the read privilege of the model parameters
or can perform model extraction techniques as demonstrated in~\cite{model_extract,juuti2019prada}.
However, it may not be possible to access the training dataset
as the DNN model is trained at cloud servers.
In addition, the test dataset may be collected in real-time 
and it becomes impossible to get enough amount of data for the precise attack.
As discussed in~\secref{sec:compare_bfa},
the attack performance of the BFA varies a lot by how the data is sampled.
To overcome such limitations, we propose {\it \underline{Ze}ro-data \underline{B}ased \underline{R}epeated bit flip
\underline{A}ttack} (ZeBRA in~\Figref{fig:intro}).

\section{Related Work}
\subsection{Adversarial Attacks on DNN Models}
Most of the studies on attacking deep learning models are based on generating adversarial examples.
Adversarial examples are inputs that are extremely difficult to distinguish by human eyes 
but successfully fool the DNN models. 
There are many prior work that try to train good adversarial examples
with imperceptible perturbations from the original images~\cite{fgsm, kurakin2017adversarial, deepfool, adv_ex_15}.
Rather than solving an optimization problem,
authors in~\cite{baluja2017adversarial} propose a neural network that transforms
an input image to an adversarial example.
Recently, adversarial weight attack has been emerged as a new domain of
the DNN attack method~\cite{bfa, tbfa, faultsneak}.
The BFA presents an iterative algorithm
that searches for bits in the parameter space that increase the DNN loss the most~\cite{bfa}.
With multiple iterations, the BFA successfully destroys the DNN model and 
makes it a random predictor.
In the BFA, however, the adversary needs to have a privilege of {\it accessing 
weights} of the victim DNN model as well as the {\it training or test dataset}
to perform the iterative bit search~\cite{bfa}.
Moreover, the same research group demonstrated a targeted BFA (T-BFA)
to make the DNN model output the same classification result on any inputs~\cite{tbfa}.
Still, it suffers from the need of the actual dataset in searching bits to be flipped.

\subsection{Flipping Bits by Physical Attack}
To make the adversarial weight attack feasible, 
there has to be a way to physically change the weight parameters stored in the memory system.
Recently, several memory fault injection techniques are developed
that threaten the the integrity of DNN models~\cite{howtoflip, laser_fault, rowhammer}.
Especially, repeated accesses to a specific row of the main memory,
i.e., dynamic RAM (DRAM), effectively cause bit flips in neighboring rows
at predictable bit locations~\cite{rowhammer}.
This cell-level attack is widely known as the row-hammer 
and its impact gets more severe as the memory technology scales down
for higher cell density~\cite{mutlu2019rowhammer}.
It is even possible to gain kernel privileges on real systems
by user-level programs 
as demonstrated by Google~\cite{google_zero}.
Moreover, a mobile system with an embedded GPU 
can be controlled by the adversary using the row-hammer attack~\cite{rowhammer_gpu}.
As the on-chip memory, i.e., static RAM (SRAM), of a mobile device has a limited capacity ($<$ 2MB),
the weight parameters are stored in the DRAM making them vulnerable to the row-hammer attack. In this work, we provide a simple yet effective method of generating synthetic data that can be utilized for the precise bit flip attack.

\section{ZeBRA: Adversarial Weight Attack with Distilled Target Data}\label{sec:zebra}
In most cases, the adversary may not have the privilege of accessing
the training dataset or a DNN could be trained over the cloud.
In addition, the test dataset may be collected in real-time by
associated sensors, e.g., cameras in self-driving cars,
or may not be easily accessed due to privacy issues, e.g., personal health records
or encrypted data.
In any of these scenarios, it is impossible to perform the BFA
on the pre-trained DNN model.
Note that the BFA requires to compute the loss by feeding in the
training or test dataset to identify the most vulnerable bits.
In this work, we propose to repeatedly generate synthetic data, 
named {\it distilled target data}, that follows the statistics 
of the pre-trained model, i.e., the mean and the standard deviation
at each batch normalization layer.
With the use of distilled target data,
the bit flip attack becomes more precise compared to the attack using a limited set of
training dataset (refer to~\secref{sec:compare_bfa}).
\subsection{Distilled Target Data}\label{sec:dist_data}
To perform the iterative bit search, we extract the synthetic
data from the DNN model itself (i.e., {\it distilled data}).
The distilled data has been presented in~\cite{zeroq}
to analyze the impact of quantization on the DNN accuracy.
However, the previous distilled data is not associated with
any target labels as it is simply used to check the KL divergence
between the model output without quantization and the one with quantization.
To analyze the bit sensitivity of the DNN model during the iterative bit search (refer to Sec.~\ref{sec:zebra_alg}), 
the synthetic data should have two properties:
i) accurately estimate the DNN loss ({\it cross entropy loss}) and ii) accurately estimate the weight gradients ({\it distilled loss}).
Thus, we assign target labels to the distilled data while generating them in the proposed {\it distilled target data}.
According to our analysis, the distilled data of a given class behaves similarly 
to the actual training dataset with the same label\footnote{More details are presented in Section B of the supplementary material.}.
Moreover, we introduce a hyper-parameter, i.e. total loss bound $\epsilon_{loss}$, 
that improves the attack performance.

In the ZeBRA, we first generate random input data $\mathbf{x}^d\in\mathbb{R}^{C\times W\times H}$ in a range of $[-1,1]$, and a random target label $\hat{y}=\{0,...,N_{C}-1\}$.
The $C$, $W$, $H$ and $N_C$ represent the number of input channels, input width, input height and number of classes, respectively.
We mini-batch the input $\mathbf{x}^d$ and obtain the one-hot encoded target label $\hat{\mathbf{y}}\in\mathbb{R}^{N_{C}}$ of the class $\hat{y}$ with the batch size $B_{distill}$.
Given the mini-batch input data $\mathbf{X}^d\in\mathbb{R}^{B_{distill}\times C\times W\times H}$ and one-hot encoded target labels $\hat{\mathbf{Y}}\in\mathbb{R}^{B_{distill}\times N_{C}}$,
a cross entropy loss is computed by \begin{equation}\label{eq:ce_loss}
\begin{split}
\mathcal{L}_{CE} = \mathcal{L}(f(\theta; \mathbf{X}^d), \hat{\mathbf{Y}}) = -\sum_{i=1}^{B_{distill}} \hat{\mathbf{Y}}_i\cdot \log(\mathbf{Y}_i)^\intercal, \ \  \mathbf{Y} = f(\theta; \mathbf{X}^d),
\end{split}
\end{equation}

where $f(\theta; \mathbf{X}^d)$ is the output probability (typically, after the softmax layer)
computed by running the DNN model with weight parameters $\theta\in\mathbb{R}^{n}$
for the given input data $\mathbf{X}^d$.
As another objective in generating the distilled target data
is to resemble the statistics of the DNN model,
a distilled loss is used as additional loss term.
The distilled loss is defined as
\begin{equation}\label{eq:dist_loss}
\mathcal{L}_{Distill} = \sum_{i=0}^{L-1} ||\widetilde{\mu_i}^d - \mu_i||_2^2 + ||\widetilde{\sigma_i}^d - \sigma_i||_2^2,
\end{equation}
where $\widetilde{\mu_i}^d$ and $\widetilde{\sigma_i}^d$ are the
average and standard deviation of feature maps at layer $i$ 
when $\mathbf{X}^d$ is fed into the DNN model with `$L$' layers.
The $\mu_i$ and $\sigma_i$ are the stored mean and standard deviation
for the $i$\textsuperscript{th} batch normalization layer.
Then, the total loss is defined as
\begin{equation}\label{eq:tot_loss}
\mathcal{L}_{total} = \lambda_{CE}\cdot \mathcal{L}_{CE} + \lambda_{Distill}\cdot \mathcal{L}_{Distill}.
\end{equation}

The generation of the distilled target data now 
becomes the optimization problem of 
finding $\mathbf{X}^d$ that minimizes $\mathcal{L}_{total}$.
A typical gradient descent is used to iteratively update
$\mathbf{X}^d$ until it reaches the total loss bound $\epsilon_{loss}$.
The $\lambda$s are used to control the strength of convergence to the model statistics
and/or the target label.
This distillation process, i.e., \texttt{distill\_target\_data}() in line 14 of \algref{alg:zebra}, 
is repeated until it generates $B_{attack}$ samples
forming an attack batch $\mathbf{X}_{attack}^d$ to be used during the iterative bit search.
Note that $B_{attack}$ is a multiple of $B_{distill}$.
Here, the definition of an attack batch is a set of data points used for selecting bits to be flipped in the ZeBRA.

\subsection{Workflow of ZeBRA Algorithm}\label{sec:zebra_alg}
The main advantage of the distilled target data is that
we no longer need either the training or test dataset for the adversarial weight attack.
More importantly, we can easily generate a new attack batch for 
the bit search process resulting in a more precise DNN weight attack.
\algref{alg:zebra} summarizes the overall process 
of the ZeBRA in selecting the well performing attack batch, i.e., a set of distilled target data,
and searching the vulnerable bits to be flipped.
It consists of two main parts: i) generating the distilled target data for the bit search process
and ii) iteratively searching bits to be flipped.

The only required data for the ZeBRA is DNN model parameters, i.e., weight ($\theta$) 
and batch normalization parameters ($\mathbf{\mu}$, $\mathbf{\sigma}$).
Then, we set the target accuracy ($A_{target}$) and the maximum number of bit flips to allow ($N_{b}^{\text{max}}$).
The algorithm generates a new set of distilled target data, $\mathbf{X}_{attack}^d$ with size $B_{attack}$,
if the attack fails to reach $A_{target}$ with bit flips less than $N_{b}^{\text{max}}$.
As the ZeBRA can generate distilled target data multiple times,
we {\it repeatedly} generate the data until the bit flip attack satisfies the attack performance.
For the evaluation of the attack performance, 
we prepare the distilled target data just for the validation, called
{\it distilled validation data}, prior to line 7 in \algref{alg:zebra}.
More details on the generation of distilled validation data,
$\mathbf{X}_{valid}^d$, will be discussed in \secref{sec:distill_valid}.

\begin{algorithm}[t]\label{alg:zebra}
\caption{ZeBRA Algorithm}
\SetAlgoVlined
\textbf{Input:} Model parameters $\theta$, $\mathbf{\mu}$, $\mathbf{\sigma}$, \\
        \hspace{8mm} Target accuracy $A_{target}$, \\
        \hspace{8mm} Attack/distill batch size $B_{attack}$, $B_{distill}$, \\
        \hspace{8mm} Maximum \# of bit flips $N_{b}^{\text{max}}$ \\
\textbf{Output:} Modified weight parameters $\theta_{attack}$, \\
        \hspace{8mm} Required \# of bit flips $N_{attack}$

 \While{$A_{target}$ $<$ $A_{attack}$}{
  \% {\it 1. Generation of distilled target data} \\
  $\mathbf{X}_{attack}^d$ = [], $\mathbf{\hat{Y}}_{attack}$ = [] \;
  $T\gets B_{attack}/B_{distill}$ \;
  \For{$i \gets 1$ to $T$}{
   Initialize input data:
   $\mathbf{X}^d\sim U(-1,1)\in\mathbb{R}^{B_{distill}\times C\times W\times H}$ \\
   Initialize one-hot encoded target labels:
   $\mathbf{\hat{Y}} \in\mathbb{R}^{B_{distill}\times N_C}$ \\
   $\mathbf{X}_{attack}^d\gets$ [$\mathbf{X}_{attack}^d|$\texttt{distill\_target \_data} ($\mathbf{X}^d$, $\mathbf{\hat{Y}}$, $\theta$, $\mathbf{\mu}$, $\mathbf{\sigma}$, $\epsilon_{loss}$)] \;
   $\mathbf{\hat{Y}}_{attack}\gets [\mathbf{\hat{Y}}_{attack}|\mathbf{\hat{Y}}]$
  }
  \% {\it 2. Iterative bit search} \\
  $N_{attack}\gets N_{b}^{\text{max}}$\;
  \For{$k \gets 1$ to $N_{b}^{\text{max}}$}{
   $\theta_{attack}\gets$\texttt{layerwise\_bit\_search}($\theta$, $\mathbf{X}_{attack}^d$, $\hat{\mathbf{Y}}_{attack}$, $\mathbf{\mu}$, $\mathbf{\sigma}$) \;
   $A_{attack}\gets f(\theta_{attack};\mathbf{X}_{valid}^d)$ \;
   $\theta\gets\theta_{attack}$ \;
   \If{$A_{target}$ $>$ $A_{attack}$}{
    $N_{attack}\gets k$\; 
    break\;
   }
  }
 }
 \Return $N_{attack}$, $\theta_{attack}$
\end{algorithm}

After the attack data $\mathbf{X}_{attack}^d$ is obtained, we can perform the iterative bit search
to identify the most vulnerable bit at each iteration.
Here, the vulnerable bit `$b_{i,l}$' at layer $l$ has the largest 
$\partial \mathcal{L}/{\partial b_{i,l}}$ where $\mathcal{L}(\cdot)$
is the cross entropy loss of a given DNN model in~\eqnref{eq:ce_loss}.
Thus, the synthesized attack data should accurately estimate the DNN loss
$\mathcal{L}$.
This is the intuition behind considering $\mathcal{L}_{CE}$ in Eq.~(\ref{eq:tot_loss}) 
when forging the attack data.
In addition, the bit sensitivity $\partial\mathcal{L}/{\partial b_{i,l}}$ can be
computed by $(\partial \mathcal{L}/{\partial\theta_l})\cdot(\partial\theta_l/{\partial b_{i,l}})$.
Thus, accurately estimating the weight gradients, 
i.e., $\partial\mathcal{L}/{\partial\theta_l}$, is important in finding vulnerable bits.
Since computing $\partial \mathcal{L}/{\partial\theta_l}$ involves multiplications
between the backpropagated gradients and input feature maps,
the attack data should approximate the statistics of feature maps at each layer.
This is why we consider $\mathcal{L}_{Distill}$ in Eq.~(\ref{eq:tot_loss}) 
when generating the attack data.

The \texttt{layerwise\_bit\_search}() in line 19 of \algref{alg:zebra}
is identical to the one presented in~\cite{bfa}.
We briefly explain the process here for the sake of completeness.
At the $k$-th iteration, as a first step, the most vulnerable bit $b_l^k$ at layer $l$ is exclusively 
selected and the inference loss $\mathcal{L}_l^k$ is computed with the bit $b_l^k$ flipped 
({\it in-layer search}).
The loss $\mathcal{L}_l^k$ is defined as
\begin{equation}\label{eq:bfa_loss_1}
\mathcal{L}_{l}^k = \mathcal{L}(f(\theta_{attack}^{k}; \mathbf{X}_{attack}^d), \hat{\mathbf{Y}}),
\end{equation}
where $\theta_{attack}^{k}$ is the weight parameter obtained by flipping the bit $b_l^k$
from $\theta_{attack}^{k-1}$.
As a second step, we identify the $j$-th layer with the maximum loss $\mathcal{L}_j^k$
and perform the permanent bit flip at $b_j^k$ ({\it cross-layer search}).
As the bit $b_j^k$ is permanently flipped, it is kept flipped at subsequent iterations.
At each bit flip attack being executed,
the post-attack accuracy ($A_{attack}$) is evaluated by using distilled validation data $\mathbf{X}_{valid}^d$.
If $A_{attack}$ is lower than $A_{target}$,
the bit search process is terminated.
At last, the modified weight parameters and the number of bit flips are returned.
If the attack fails, the ZeBRA repeats generating the new attack batch
(thus, named {\it Zero-data Based Repeated bit flip Attack}).
\vspace{-2mm}
\section{Attack Performance of ZeBRA}\label{sec:exp_result}
In this section, we compare the attack performance of ZeBRA 
to BFA on various DNN models\footnote{The code for the ZeBRA will be available at https://github.com/pdh930105/ZeBRA.}.
The novelty of the proposed ZeBRA algorithm is in that
we do not need either training or test dataset, unlike the BFA.
To allow the BFA to work, we assume that the BFA can sample 
a mini-batch from the actual dataset, e.g., $B_{attack}=64$,
for the iterative bit search.

\subsection{Experimental Setup}\label{sec:exp_setup}
\noindent\textbf{Datasets and DNN Models:} As test benchmarks, we select the two well-known 
image classification datasets: CIFAR-10~\cite{cifar10} and ImageNet~\cite{imagenet}.
The CIFAR-10 has 10 different object classes while ImageNet has 1,000 different classes.
Each dataset is divided into training, validation and test datasets.
The BFA (baseline) in our experiments samples a mini-batch from 
the training dataset for the bit search process (it may not be possible in the real-world scenario).
Note that the ZeBRA generates its own attack batches prior to the bit search process.
The runtime overhead of generating the attack data is discussed in \secref{sec:runtime}.
For CIFAR-10 dataset, four different ResNet models (ResNet-20/32/44/56) are used
for evaluating the attack performance of the ZeBRA~\cite{resnet}.
For ImageNet dataset, VGG11, Inception-v3, ResNet-18/34/50 and MobileNetV2 are used for the evaluation~\cite{vgg, inception, resnet,mobilenetv2}.
All the experiments are conducted on NVIDIA GeForce RTX 2080Ti (11GB memory).
\vspace{1mm}

\noindent\textbf{Selection of ZeBRA Hyper-parameters:} There are several hyper-parameters
to be determined to perform an effective bit flip attack with the ZeBRA.
They are mini-batch size $B_{distill}$, coefficients $\lambda_{CE}$ and $\lambda_{Distill}$ 
in \eqnref{eq:tot_loss}, and total loss bound $\epsilon_{loss}$.
We fix $B_{attack}$ of the ZeBRA to 64 that matches the mini-batch size of the BFA.
To verify the sole impact of the distilled target data $\mathbf{X}_{attack}^d$
and its associated hyper-parameters, 
the distilled validation data $\mathbf{X}_{valid}^d$ in line 20 of
\algref{alg:zebra} is replaced with the actual validation
dataset $\mathbf{X}_{valid}$ of CIFAR-10 or ImageNet.
The impact of using the distilled validation data instead of the actual validation
dataset will be discussed in~\secref{sec:distill_valid}.

To select the optimal hyper-parameters for generating distilled target data,
we tested different hyper-parameters on ResNet-20 with CIFAR-10 dataset.
The ResNet-20 model is quantized to 8bit\footnote{The quantized DNN model 
is more robust to bit flips than the one in floating-point representation~\cite{fault_inj, bfa}.}.
Tests on other network architectures, quantization levels (6bit and 4bit), or dataset (ImageNet) 
show similar trends as shown in~\Tabref{tab:hp_select}.
We generated 40 different sets of $\mathbf{X}_{attack}^d$ for each hyper-parameter combination 
\{$\lambda_{CE}$, $\lambda_{Distill}$, $\epsilon_{loss}$\}
and obtained the minimum number of bit flips to achieve $A_{target}=10$\% (i.e., making a random predictor).
As a result, we select $B_{distill}=16$, 
$B_{attack}=64$, $\lambda_{CE}=0.2$, $\lambda_{Distill}=0.1$, and $\epsilon_{loss}=10$
when generating $\mathbf{X}_{attack}^d$ for the rest of our experiments.
Note that the ZeBRA fails to attack the model with random data ($\lambda_{CE}=0$ and $\lambda_{Distill}=0$)
proving that the distilled target data is definitely required to perform an effective attack.

\begin{table*}[t]
\centering
\caption{The attack performance of ZeBRA at different hyper-parameter combinations of ($\lambda_{CE}$, $\lambda_{Distill}$, $\epsilon_{loss}$)
    on ResNet-20 with CIFAR-10 dataset:
    The ZeBRA outputs the minimum number of bit flips, i.e., only 8 bits, with $\lambda_{CE}=0.2$, $\lambda_{Distill}=0.1$, and $\epsilon_{loss}=10$}\vspace{2mm}
\label{tab:hp_select}
\scalebox{0.50}{
\begin{tabular}{ccccccccccccccccc}
\specialrule{.2em}{.1em}{.1em} 
\multicolumn{2}{c}{\multirow{2}{*}{\textbf{\begin{tabular}[c]{@{}c@{}}ResNet-20\\ (CIFAR-10)\end{tabular}}}}   & \multicolumn{3}{c}{$\lambda_{CE}$ = \textbf{0}} & \multicolumn{3}{c}{$\lambda_{CE}$ = \textbf{0.1}}             & \multicolumn{3}{c}{$\lambda_{CE}$ = \textbf{0.2}}                  & \multicolumn{3}{c}{$\lambda_{CE}$ = \textbf{0.5}}        & \multicolumn{3}{c}{$\lambda_{CE}$ = \textbf{1.0}}        \\ \cline{3-17} 
\multicolumn{2}{c}{}                                                                                           & $\epsilon_{loss}$ = \textbf{1} & \textbf{10}      & \textbf{100} & \textbf{1} & \textbf{10}           & \textbf{100} & \textbf{1} & \textbf{10} & \textbf{100} & \textbf{1} & \textbf{10} & \textbf{100} & \textbf{1} & \textbf{10} & \textbf{100} \\ \hline\hline
\multirow{5}{*}{\textbf{\begin{tabular}[c]{@{}c@{}}Bit Flips\\ (Best\\/ Mean)\end{tabular}}} 

                                & 
                                $\lambda_{Distill}$ = \textbf{0} & 40 / 40    & 40 / 40 &   40 / 40 &   40 / 40 &   40 / 40 &   40 / 40 & 18 / 25      & 40 / 40    & 40 / 40     & 8 / 22      & 40 / 40    & 40 / 40     & 8 / 23  &   40 / 40 &   40 / 40      \\

                                & 
                                $\lambda_{Distill}$ = \textbf{0.1} & 13 / 31    & 10 / 13 & 40 / 40      & 15 / 29    & \textbf{9 / 11} & 40 / 40      & 17 / 30    & \underline{\textbf{8 / 12 }}     & 40 / 40      & 15 / 28    & 10 / 28     & 40 / 40 &  13 / 27 & 12 / 26 & 40 / 40      \\
                                                                                                
                                                                    & 
                                  $\lambda_{Distill}$ = \textbf{0.2} & 15 / 34    & 9 / 24          & 40 / 40      & 17 / 33    & 16 / 24               & 40 / 40      & 24 / 36    & 15 / 26     & 40 / 40      & 14 / 35    & 21 / 29     & 40 / 40 &  10 / 35  & 23 / 30  &   40 / 40      \\
                                                                                                & $\lambda_{Distill}$ = \textbf{0.5} & 21 /38     & 18 / 30          & 40 / 40      & 17 / 37    & 15 / 28               & 40 / 40      & 26 / 38    & 14 / 27     & 40 / 40    & 21 / 37  & 9 / 27    & 40 / 40     & 19 / 37 & 15 / 26 & 28 / 39      \\
                                                                                                & $\lambda_{Distill}$ = \textbf{1.0} & 20 / 37    & 16 / 30          & 10 / 15      & 19 / 33    & 13 / 31               & 11 / 12      & 22 / 37    & 16 / 33     & 10 / 13      & 22 / 37    & 14 / 29     &  \textbf{10 / 11} & 17 / 35 & 13 / 29 &  \textbf{9 / 11}      \\ 
\specialrule{.2em}{.1em}{.1em} 
\end{tabular}}
\end{table*}

\subsection{Comparison to BFA}\label{sec:compare_bfa}

\subsubsection{Comparison on CIFAR-10}
A random predictor has classification accuracy of 10\%, due to CIFAR-10 dataset has 10 classes, which is target accuracy for each attack method.
Four different ResNet models (ResNet-20/32/44/56) are selected as benchmarks
to evaluate the attack performance.
The ResNet models at various quatization levels (8bit, 6bit and 4bit) are tested.
\Tabref{tab:compare_cifar10} summarizes the attack performance
of the BFA and the proposed ZeBRA on CIFAR-10 dataset.

The both BFA and ZeBRA are performed 50 times with different seed values.
Again, it is challenging for the BFA to obtain 50 different mini-batches from the actual dataset 
while the ZeBRA can easily self-generate any number of mini-batches.
As our experimental result shows it is not guaranteed for the BFA to obtain the minimum number of bit flips,
i.e., 8.58 bits on average, without accessing the large amount of training dataset.
The mean and standard deviation of the required number of bit flips 
to fully destroy a given DNN model are large for the BFA:
30.3/15.4, 23.5/12.2, 30.8/13.3, and 22.4/10.5 for ResNet-20, ResNet-32, ResNet-44, and ResNet-56 (8bit), respectively.
Similar statistics are observed for 6bit and 4bit quantized models as provided in \Tabref{tab:compare_cifar10}.
In addition, 26\% of trials on average failed in attacking the DNN model ($N_{b}^{\text{max}}$ is set to 50).
This implies that the selection of a mini-batch for the BFA significantly impacts the attack performance.

On the contrary, the ZeBRA has freedom in generating the attack dataset.
The attack performance for 50 trials of the ZeBRA is summarized in~\Tabref{tab:compare_cifar10}.
With the ZeBRA, the minimum number of bit flips to completely destroy the DNN model
is 9.50 bits on average (0.92 bits higher than the BFA).
However, in the ZeBRA, it is guaranteed to achieve the minimum number of bit flips 
as we can examine the model as much as we can before physically attacking the DNN model, e.g., via row-hammering.
In addition, none of the trials failed which implies that the ZeBRA is more reliable. 
The mean and standard deviation of the required number of bit flips are
2.0$\times$ and 8.3$\times$ smaller than the BFA on average.


\begin{table*}[t]
\centering
\caption{The comparison of the attack performance between the BFA and the proposed ZeBRA on CIFAR-10 dataset}\vspace{2mm}
\label{tab:compare_cifar10}
\scalebox{0.6}{
\begin{tabular}{ccccccccccc}
\specialrule{.2em}{.1em}{.1em}
\multirow{2}{*}{\textbf{Model}}     & \multirow{2}{*}{\textbf{\begin{tabular}[c]{@{}c@{}}Quant.\\ Level \\($N_Q$)\end{tabular}}} & \multirow{2}{*}{\textbf{\begin{tabular}[c]{@{}c@{}}Original \\ Accuracy\\ (Top-1 {[}\%{]})\end{tabular}}} & \multicolumn{4}{c}{\textbf{BFA}}                                                                                                                                                                                                                                                   & \multicolumn{4}{c}{\textbf{ZeBRA}}                                                                                                                                                                                                                                                  \\ \cline{4-11} 
                                    &                                                                                  &                                                                                                             & \textbf{\begin{tabular}[c]{@{}c@{}}Bit Flips\\ (Best)\end{tabular}} & \textbf{\begin{tabular}[c]{@{}c@{}}Bit Flips\\ (Worst)\end{tabular}} & \textbf{\begin{tabular}[c]{@{}c@{}}Mean\\ / Stdev\end{tabular}} & \textbf{\begin{tabular}[c]{@{}c@{}}Avg. Accuracy\\After Attack\end{tabular}} & \textbf{\begin{tabular}[c]{@{}c@{}}Bit Flips\\ (Best)\end{tabular}} & \textbf{\begin{tabular}[c]{@{}c@{}}Bit Flips\\ (Worst)\end{tabular}} & \textbf{\begin{tabular}[c]{@{}c@{}}Mean\\ / Stdev\end{tabular}} & \textbf{\begin{tabular}[c]{@{}c@{}}Avg. Accuracy\\After Attack\end{tabular}} \\ \hline\hline
\multirow{3}{*}{\textbf{ResNet-20}} & \textbf{8bit}                                                                    & 92.41                                                                                                       & 8                                                                   & 50 (fail)                                                            & 30.3 / 15.4                                                     & 10.05                                                               & 8                                                                   & 11                                                                   & 10.5 / 0.9                                                      & 10.00                                                                \\
                                    & \textbf{6bit}                                                                    & 92.18                                                                                                       & 7                                                                   & 50 (fail)                                                            & 24.6 / 17.6                                                     & 10.07                                                               & 9                                                                   & 10                                                                   & 9.8 / 0.4                                                       & 10.00                                                                \\
                                    & \textbf{4bit}                                                                    & 87.59                                                                                                       & 6                                                                   & 50 (fail)                                                            & 16.6 / 11.1                                                     & 9.99                                                                & 9                                                                   & 15                                                                   & 14.4 / 1.6                                                      & 9.77                                                                 \\ \hline
\multirow{3}{*}{\textbf{ResNet-32}} & \textbf{8bit}                                                                    & 92.77                                                                                                       & 7                                                                   & 50 (fail)                                                            & 23.5 / 12.2                                                     & 10.08                                                               & 9                                                                   & 13                                                                   & 10.6 / 0.8                                                      & 10.00                                                                \\
                                    & \textbf{6bit}                                                                    & 92.55                                                                                                       & 10                                                                  & 50 (fail)                                                            & 22.4 / 10.2                                                     & 10.05                                                               & 9                                                                   & 13                                                                   & 9.8 / 1.2                                                       & 10.00                                                                \\
                                    & \textbf{4bit}                                                                    & 92.20                                                                                                       & 10                                                                  & 50 (fail)                                                            & 18.7 / 9.6                                                      & 10.01                                                               & 10                                                                  & 19                                                                   & 14.5 / 3.0                                                      & 9.96                                                                 \\ \hline
\multirow{3}{*}{\textbf{ResNet-44}} & \textbf{8bit}                                                                    & 93.34                                                                                                       & 11                                                                  & 50 (fail)                                                            & 30.8 / 13.3                                                     & 10.07                                                               & 14                                                                  & 20                                                                   & 18.5 / 1.4                                                      & 10.00                                                                \\
                                    & \textbf{6bit}                                                                    & 93.08                                                                                                       & 9                                                                   & 50 (fail)                                                            & 26.2 / 12.2                                                     & 10.04                                                               & 11                                                                  & 18                                                                   & 14.9 / 1.9                                                      & 10.00                                                                \\
                                    & \textbf{4bit}                                                                    & 87.83                                                                                                       & 9                                                                   & 50 (fail)                                                            & 26.7 / 14.7                                                     & 10.08                                                               & 9                                                                   & 16                                                                   & 13.5 / 1.5                                                      & 10.00                                                                \\ \hline
\multirow{3}{*}{\textbf{ResNet-56}} & \textbf{8bit}                                                                    & 93.50                                                                                                       & 8                                                                   & 50 (fail)                                                            & 22.4 / 10.5                                                     & 10.01                                                               & 9                                                                   & 18                                                                   & 13.4 / 2.3                                                      & 10.00                                                                \\
                                    & \textbf{6bit}                                                                    & 93.32                                                                                                       & 10                                                                  & 50 (fail)                                                            & 27.2 / 13.2                                                     & 10.01                                                               & 7                                                                   & 17                                                                   & 13.2 / 1.9                                                      & 10.00                                                                \\
                                    & \textbf{4bit}                                                                    & 89.61                                                                                                       & 8                                                                   & 50 (fail)                                                            & 46.0 / 11.1                                                     & 10.70                                                               & 10                                                                  & 16                                                                   & 13.8 / 1.3                                                      & 10.00                                                                \\ \specialrule{.2em}{.1em}{.1em}
 \end{tabular}}
\end{table*}

\subsubsection{Comparison on ImageNet}
To generalize the effectiveness of the ZeBRA,
we also compared the attack performance on ImageNet dataset.
Note that a random predictor for ImageNet has classification accuracy of 0.1\% (Top-1) 
and we set the target accuracy to 0.2\% which is identical to the prior work~\cite{bfa}.
For the evaluation, we select VGG11, Inception-v3, three different ResNet models (ResNet-18/34/50), and
a mobile-friendly DNN model (MobileNetV2), quantized at 8bit.
\Tabref{tab:compare_imagenet} summarizes the attack performance of the BFA
and the ZeBRA on ImageNet dataset.

Similarly, 50 trials with different seeds are performed for both the BFA and ZeBRA.
As our experimental results show, 
DNN models trained on ImageNet are more susceptible to the adversarial weight attack.
It requires less than 6 bits to completely destroy the DNN model.
We conjecture that decision boundaries are close to each other
for the DNN model on ImageNet as it needs to partition the feature space into 1,000 different regions.
Thus, a slight modification to decision boundaries significantly impacts the accuracy.
The minimum number of bit flips to destroy the DNN model
is 3.83 bits for BFA and 4 bits for ZeBRA on average.
Note that the ZeBRA guarantees finding the minimum number of bit flips 
without accessing the actual dataset.
The mean and standard deviation of the required number of bit flips are
1.6$\times$ and 3.5$\times$ smaller than the BFA on average ({\it thus, ZeBRA is more precise}).

\begin{table*}[t]
\centering
\caption{The comparison of the attack performance between the BFA and the proposed ZeBRA on ImageNet dataset}\vspace{2mm}
\label{tab:compare_imagenet}
\scalebox{0.6}{
\begin{tabular}{ccccccccccc}
\specialrule{.2em}{.1em}{.1em}
\multirow{2}{*}{\textbf{Model}}       & \multirow{2}{*}{\textbf{Accuracy}} & \multirow{2}{*}{\textbf{\begin{tabular}[c]{@{}c@{}}Original\\ Accuracy\\ {[}\%{]}\end{tabular}}} & \multicolumn{4}{c}{\textbf{BFA}}                                                                                                                                                                                                                                                                      & \multicolumn{4}{c}{\textbf{ZeBRA}}                                                                                                                                                                                                                                                                    \\ \cline{4-11} 
                                      &                                    &                                                                                                & \textbf{\begin{tabular}[c]{@{}c@{}}Bit Flips\\ (Best)\end{tabular}} & \textbf{\begin{tabular}[c]{@{}c@{}}Bit Flips\\ (Worst)\end{tabular}} & \textbf{\begin{tabular}[c]{@{}c@{}}Mean\\/ Stdev\end{tabular}}          & \textbf{\begin{tabular}[c]{@{}c@{}}Avg. Accuracy \\ After Attack\end{tabular}} & \textbf{\begin{tabular}[c]{@{}c@{}}Bit Flips\\ (Best)\end{tabular}} & \textbf{\begin{tabular}[c]{@{}c@{}}Bit Flips\\ (Worst)\end{tabular}} & \textbf{\begin{tabular}[c]{@{}c@{}}Mean\\/ Stdev\end{tabular}}          & \textbf{\begin{tabular}[c]{@{}c@{}}Avg. Accuracy \\ After Attack\end{tabular}} \\ \hline\hline

\multirow{2}{*}{\textbf{VGG11}}   & \textbf{Top-1}                     & 70.24                                                                                          & \multirow{2}{*}{9}                                                  & \multirow{2}{*}{35}                                                  & \multirow{2}{*}{\begin{tabular}[c]{@{}c@{}}16.48\\/ 6.46\end{tabular}}  & 0.18                                                                           & \multirow{2}{*}{6}                                                  & \multirow{2}{*}{10}                                                   & \multirow{2}{*}{\begin{tabular}[c]{@{}c@{}}8.48\\/ 0.96 \end{tabular}}  & 0.15                                                                           \\
                                      & \textbf{Top-5}                     & 89.68                                                                                          &                                                                     &                                                                      &                                                                         & 0.77                                                                           &                                                                     &                                                                      &                                                                         & 0.71                                                                           \\ \hline

\multirow{2}{*}{\textbf{Inception-v3}}   & \textbf{Top-1}                     & 76.85                                                                                          & \multirow{2}{*}{2}                                                  & \multirow{2}{*}{6}                                                  & \multirow{2}{*}{\begin{tabular}[c]{@{}c@{}}3.21\\/ 1.06\end{tabular}}  & 0.12                                                                           & \multirow{2}{*}{2}                                                  & \multirow{2}{*}{4}                                                   & \multirow{2}{*}{\begin{tabular}[c]{@{}c@{}}2.5\\/ 0.87 \end{tabular}}  & 0.15                                                                           \\
                                      & \textbf{Top-5}                     & 93.33                                                                                          &                                                                     &                                                                      &                                                                         & 3.45                                                                           &                                                                     &                                                                      &                                                                         & 1.73                                                                           \\ \hline

\multirow{2}{*}{\textbf{ResNet-18}}   & \textbf{Top-1}                     & 69.50                                                                                          & \multirow{2}{*}{5}                                                  & \multirow{2}{*}{15}                                                  & \multirow{2}{*}{\begin{tabular}[c]{@{}c@{}}8.12\\/ 1.76\end{tabular}}  & 0.15                                                                           & \multirow{2}{*}{6}                                                  & \multirow{2}{*}{9}                                                   & \multirow{2}{*}{\begin{tabular}[c]{@{}c@{}}7.21\\/ 1.35\end{tabular}}  & 0.15                                                                           \\
                                      & \textbf{Top-5}                     & 88.97                                                                                          &                                                                     &                                                                      &                                                                         & 1.52                                                                           &                                                                     &                                                                      &                                                                         & 1.69                                                                           \\ \hline
\multirow{2}{*}{\textbf{ResNet-34}}   & \textbf{Top-1}                     & 73.13                                                                                          & \multirow{2}{*}{4}                                                  & \multirow{2}{*}{17}                                                  & \multirow{2}{*}{\begin{tabular}[c]{@{}c@{}}9.53\\/ 2.81\end{tabular}}  & 0.15                                                                           & \multirow{2}{*}{5}                                                  & \multirow{2}{*}{6}                                                   & \multirow{2}{*}{\begin{tabular}[c]{@{}c@{}}5.07\\/ 0.26\end{tabular}}  & 0.14                                                                           \\
                                      & \textbf{Top-5}                     & 91.38                                                                                          &                                                                     &                                                                      &                                                                         & 1.87                                                                           &                                                                     &                                                                      &                                                                         & 2.56                                                                           \\ \hline
\multirow{2}{*}{\textbf{ResNet-50}}   & \textbf{Top-1}                     & 75.84                                                                                          & \multirow{2}{*}{2}                                                  & \multirow{2}{*}{30}                                                  & \multirow{2}{*}{\begin{tabular}[c]{@{}c@{}}8.42\\/ 3.32\end{tabular}}  & 0.14                                                                           & \multirow{2}{*}{4}                                                  & \multirow{2}{*}{6}                                                   & \multirow{2}{*}{\begin{tabular}[c]{@{}c@{}}4.70\\/ 0.92\end{tabular}}  & 0.15                                                                           \\
                                      & \textbf{Top-5}                     & 92.81                                                                                          &                                                                     &                                                                      &                                                                         & 0.80                                                                           &                                                                     &                                                                      &                                                                         & 1.05                                                                           \\ \hline
\multirow{2}{*}{\textbf{MobileNetV2}} & \textbf{Top-1}                     & 71.14                                                                                          & \multirow{2}{*}{1}                                                  & \multirow{2}{*}{8}                                                   & \multirow{2}{*}{\begin{tabular}[c]{@{}c@{}}2.65\\/ 0.031\end{tabular}} & 0.14                                                                           & \multirow{2}{*}{1}                                                  & \multirow{2}{*}{2}                                                   & \multirow{2}{*}{\begin{tabular}[c]{@{}c@{}}1.68\\/ 0.014\end{tabular}} & 0.12                                                                           \\
                                      & \textbf{Top-5}                     & 90.01                                                                                          &                                                                     &                                                                      &                                                                         & 0.66                                                                           &                                                                     &                                                                      &                                                                         & 0.61                                                                           \\ \specialrule{.2em}{.1em}{.1em}
\end{tabular}}
\end{table*}

\begin{figure}[t]
\centering
\begin{minipage}[b]{0.45\linewidth}
\centering
\includegraphics[height=4.2cm]{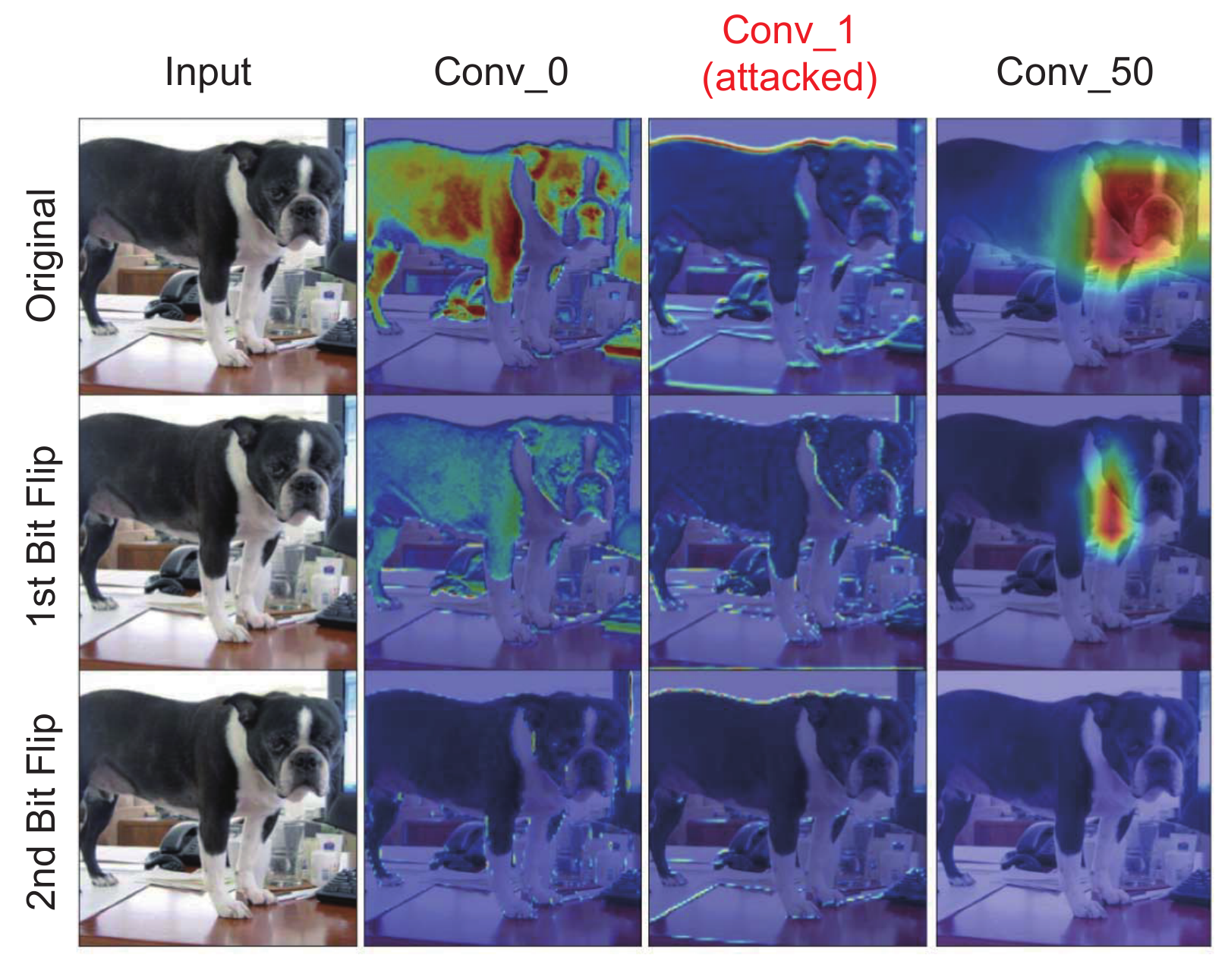}
\caption{The location of the attention map~\cite{gradcam} significantly changes 
    by only a couple of bit flips in MobileNetV2.}
\label{fig:gradcam}
\end{minipage}
\quad
\begin{minipage}[b]{0.45\linewidth}
\centering
\includegraphics[height=4.2cm]{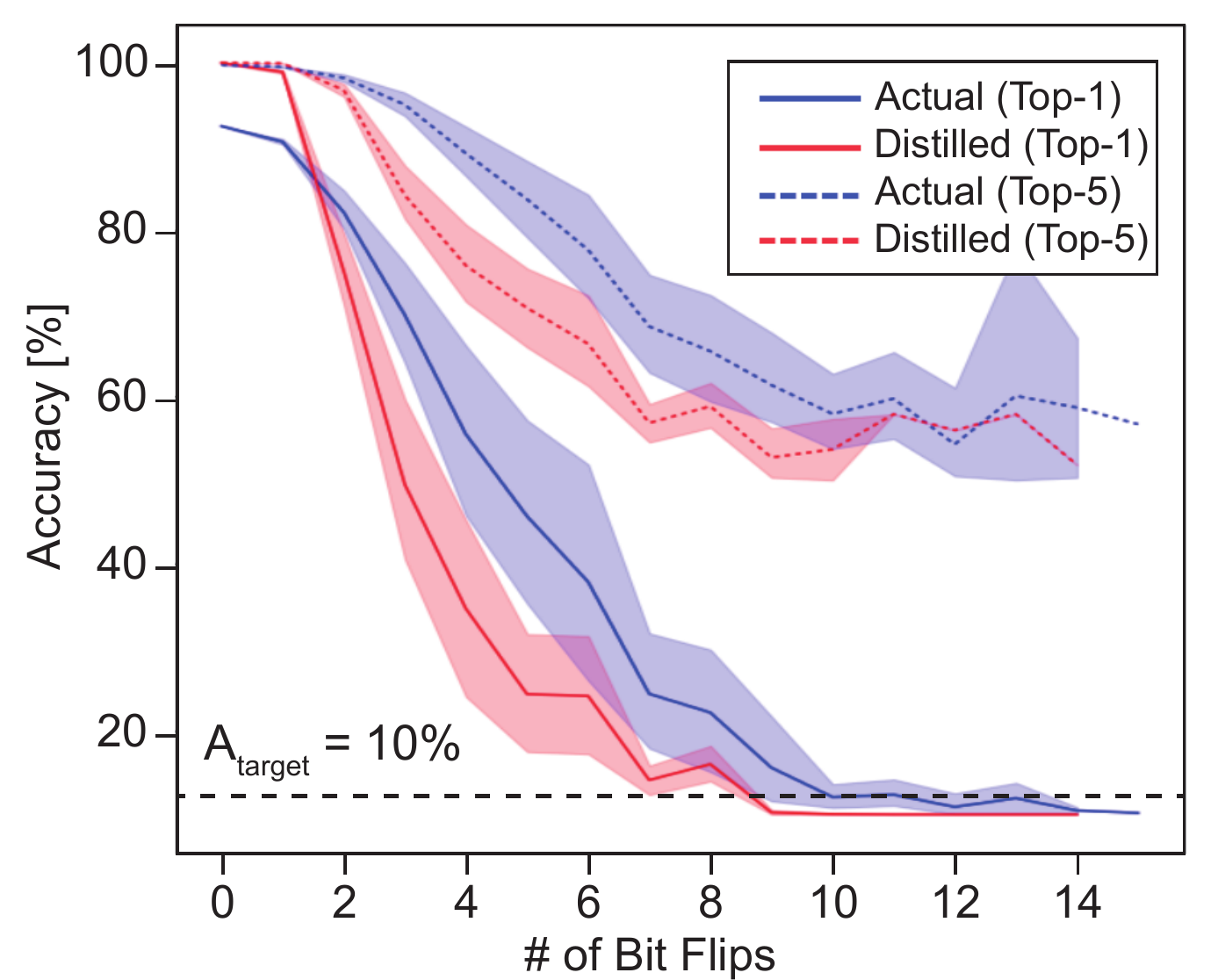}
\caption{The accuracy evaluated by the actual dataset (blue) and 
    the distilled validation data (red) with 20 ZeBRA trials.}
\label{fig:acc_curve}
\end{minipage}
\end{figure}

A noticeable result is that even 1 bit is enough to change MobileNetV2 into a random predictor.
\Figref{fig:gradcam} shows the attention maps at several convolution layers in MobileNetV2 extracted by Grad-CAM~\cite{gradcam}.
A significant weight change in the depthwise convolution layer makes a single output channel to have large values (either positive or negative).
Mostly, mobile-friendly DNNs have a depthwise convolution layer followed by a pointwise (1$\times$1) convolution layer to reduce the number of computations\footnote{More experimental results on mobile-friendly DNNs are presented in Section C of the supplementary material.}.
Thus, the large-valued feature map impacts all output channels after the 1$\times$1 convolutions.
This has a huge impact on the security of efficient DNNs~\cite{efficientnet, hw_nas},
as they are more fragile to adversarial weight attacks.

\subsection{ZeBRA with Distilled Validation Data}\label{sec:distill_valid}

So far, we evaluated the attack performance, i.e., line 20 in \algref{alg:zebra}, 
using the actual validation dataset for both the BFA and ZeBRA.
This is because we first have to verify that the distilled target data 
works well as the attack data $\mathbf{X}_{attack}^d$.
However, a genuine zero-data based bit flip attack is realized 
when we can evaluate the accuracy with the distilled target data as well.
We call this synthesized data for accuracy evaluation as {\it distilled validation data} $\mathbf{X}_{valid}^d$.
The $\mathbf{X}_{valid}^d$ is generated once prior to line 7 in \algref{alg:zebra}
with the same procedure from line 12 to 15 in \algref{alg:zebra}.
For CIFAR-10 and ImageNet, 3.2k and 10k images are self-generated.
The changes in Top-1 and Top-5 accuracy after each bit filp attack with 20 ZeBRA trials on ResNet-20 with CIFAR-10 
are provided in \Figref{fig:acc_curve}.
The blue curve is the accuracy when the attacked model $\theta_{attack}$ is evaluated with the actual validation dataset.
The red curve shows the accuracy when $\mathbf{X}_{valid}^d$ is used instead.
As expected, there are some gaps as it is extremely difficult to exactly match the accuracy with the actual dataset.
However, the trend of accuracy drop evaluated by $\mathbf{X}_{valid}^d$ follows well with the one
evaluated by the actual dataset.


For better fidelity of the ZeBRA, 
we add Top-5 target accuracy (e.g., 52\% for CIFAR-10 and 1\% for ImageNet) 
as it is another good measure to check whether the model became a random predictor or not.
As the estimated accuracy with $\mathbf{X}_{valid}^d$ drops faster than
the actual accuracy, the resulting number of bit flips on average by the ZeBRA (reported in \Tabref{tab:distill_val})
reduces by 2.4 bits for CIFAR-10 when compared to the result in \Tabref{tab:compare_cifar10}.
Similar number of bit flips on average is observed for ImageNet 
when compared to the result in \Tabref{tab:compare_imagenet}.
The average accuracy after attack, however, is higher due to the error in accuracy estimation 
(11.3$\sim$15.5\%, not 10\% for CIFAR-10 and 1.1$\sim$12.3\%, not 0.2\% for ImageNet).
Still, the minimum Top-1 accuracy near the target accuracy was achieved.
Thus, we can say that DNN models are completely destroyed with the ZeBRA
without accessing the actual dataset.


\begin{table*}[t]
\centering
\caption{The attack performance of ZeBRA with distilled validation dataset on all benchmarks for CIFAR-10 and ImageNet}\vspace{2mm}
\label{tab:distill_val}
\scalebox{0.6}{
\begin{tabular}{ccccc||ccccc}
\specialrule{.2em}{.1em}{.1em}
\multirow{2}{*}{\textbf{Model}} & \multicolumn{4}{c||}{\textbf{ZeBRA w/ Distilled Validation Data (CIFAR-10)}}                                                                                                                                                                                                           & \multirow{2}{*}{\textbf{Model}} & \multicolumn{4}{c}{\textbf{ZeBRA w/ Distilled Validation Data (ImageNet)}}                                                                                                                                                                                                            \\ \cline{2-5} \cline{7-10} 
                                & \textbf{\begin{tabular}[c]{@{}c@{}}Bit Flips\\ (Best)\end{tabular}} & \textbf{\begin{tabular}[c]{@{}c@{}}Bit Flips\\ (Worst)\end{tabular}} & \textbf{\begin{tabular}[c]{@{}c@{}}Mean\\ / Stdev\end{tabular}} & \textbf{\begin{tabular}[c]{@{}c@{}}Top-1 Accuracy\\ (Min / Avg) \end{tabular}} &                                 & \textbf{\begin{tabular}[c]{@{}c@{}}Bit Flips\\ (Best)\end{tabular}} & \textbf{\begin{tabular}[c]{@{}c@{}}Bit Flips\\ (Worst)\end{tabular}} & \textbf{\begin{tabular}[c]{@{}c@{}}Mean\\ / Stdev\end{tabular}} & \textbf{\begin{tabular}[c]{@{}c@{}}Top-1 Accuracy\\ (Min / Avg) \end{tabular}} \\ \hline\hline
\textbf{ResNet-20}              & 7                                                                   & 10                                                                   & 8.5 / 1.4                                                       & 9.9 / 11.3                                                                  & \textbf{ResNet-18}              & 4                                                                   & 10                                                                   & 5.9 / 1.8                                                       & 0.3 / 8.4                                                                   \\
\textbf{ResNet-32}              & 6                                                                   & 10                                                                   & 7.2 / 1.6                                                       & 10.7 / 13.0                                                                  & \textbf{ResNet-34}              & 2                                                                   & 10                                                                    & 3.4 / 1.7                                                       & 0.4 / 8.4                                                                   \\
\textbf{ResNet-44}              & 13                                                                  & 19                                                                   & 16.3 / 1.4                                                      & 10.1 / 15.5                                                                  & \textbf{Resnet-50}              & 3                                                                   & 7                                                                    & 4.4 / 1.1                                                       & 0.1 / 12.3                                                                   \\
\textbf{ResNet-56}              & 8                                                                   & 16                                                                   & 11.6 / 2.6                                                      & 10.0 / 15.2                                                                   & \textbf{MobileNetV2}            & 1                                                                   & 4                                                                    & 2.4 / 0.8                                                       & 0.1 / 1.1                                                                   \\ \specialrule{.2em}{.1em}{.1em}
\end{tabular}}\vspace{-2mm}
\end{table*}

\subsection{Runtime of ZeBRA Algorithm}\label{sec:runtime}

As the ZeBRA requires to distill the attack batch prior to the iterative bit search,
we analyze the runtime overhead compared to the BFA.
The runtime for each stage during the BFA or ZeBRA is reported in \tabref{tab:runtime}.
As the number of searched bits differs by the DNN models and datasets,
we multiply the average number of bit flips on each benchmark and the runtime for a single bit search process.
The average bit flips reported in \tabref{tab:compare_cifar10} for CIFAR-10 and \tabref{tab:compare_imagenet} for ImageNet are used for the BFA.
The average bit flips reported in \tabref{tab:distill_val} are used for the ZeBRA utilizing the distilled validation data for its accuracy evaluation.
Due to the additional data generation process of the ZeBRA,
it takes 1.03$\times$ and 4.18$\times$ longer to perform the iterative bit search.
This runtime overhead is insignificant since the iterative bit search is performed offline 
prior to the physical bit flip attack.
Thus, the adversary has little runtime constraint on searching bits to be flipped as well as 
generating attack data $\mathbf{X}_{attack}^d$.
As emphasized by this work, the attack becomes more effective as training/test datasets
are no longer needed to perform the adversarial weight attack with the proposed method.

\begin{table}[]
\centering
\caption{The runtime comparison between the BFA (only bit search) and ZeBRA (data generation + bit search) on various benchmarks}\vspace{2mm}
\label{tab:runtime}
\scalebox{0.62}{
\begin{tabular}{cccc||cccc}
\specialrule{.2em}{.1em}{.1em}
\multirow{2}{*}{\textbf{CIFAR-10}} & \textbf{BFA}            & \multicolumn{2}{c||}{\textbf{ZeBRA with $\mathbf{X}_{valid}^d$}}                 & \multirow{2}{*}{\textbf{ImageNet}} & \textbf{BFA}            & \multicolumn{2}{c}{\textbf{ZeBRA with $\mathbf{X}_{valid}^d$}}                 \\ \cline{2-4} \cline{6-8} 
                                   & \textbf{Bit Search (s)} & \textbf{Distill Data (s)} & \textbf{Bit Search (s)} &                                    & \textbf{Bit Search (s)} & \textbf{Distill Data (s)} & \textbf{Bit Search (s)} \\ \hline\hline
\textbf{ResNet-20}                 & 2.03           & 2.09             & 0.57           & \textbf{ResNet-18}                 & {5.03}           & {4.03}             & {3.66}           \\ 
\textbf{ResNet-32}                 & 3.53           & 2.36             & 1.08           & \textbf{ResNet-34}                 & {15.53}          & {40.07}            & {5.54}           \\ 
\textbf{ResNet-44}                 & 8.32           & 3.26             & 4.40           & \textbf{ResNet-50}                 & {21.47}          & {13.10}            & {11.22}          \\ 
\textbf{ResNet-56}                 & 9.41           & 3.89             & 4.87           & \textbf{MobileNetV2}               & {2.09}           & {21.36}            & {1.90}           \\ \specialrule{.2em}{.1em}{.1em}
\end{tabular}}
\end{table}

\section{Conclusion}

We proposed a zero-data based repeated bit flip attack
having the ability of generating its own attack and validation data
to perform the bit flip attack.
As the adversary requires less knowledge for the attack,
the ZeBRA will become a significant threat to any safety-critical deep learning applications.
Especially, as demonstrated by this work, mobile-friendly DNN models require
more attention for the improved robustness to adversarial weight attacks.
In terms of the attack performance of the ZeBRA,
a better way of generating distilled validation data needs to be developed
to improve the quality of accuracy estimation which remains as our future work.
To improve the quality of the distilled validation data,
we may collect several data samples per target label and apply deep metric learning
to better cluster the synthesized samples of each target label.
Moreover, it will be useful to extend the study on other important tasks 
such as semantic segmentation and language modeling.

\section*{Acknowledgment}
This work was supported in part by Samsung Research Funding Incubation Center of Samsung Electronics (SRFC-IT1902-03), the National Research Foundation of Korea (NRF) under Grant NRF-2019R1G1A1008751, and the Institute of Information and Communications Technology Planning and Evaluation (IITP) Grant funded by the Korean Government (MSIT) (IITP-2021-2018-0-01433; ITRC support program, and No. 2019-0-00533; Research on CPU vulnerability detection and validation). We also thank Dr. Deliang Fan and Adnan Siraj Rakin for their valuable inputs.

\bibliography{egbib}

\end{document}


\maketitle

\begin{alphasection}
\section{Attack Performance of ZeBRA at Different Batch Sizes}
\label{batch_size performance}
\tabref{tab:zebra_bsize} summarizes the impact of the mini-batch size $B_{distill}$
on the ZeBRA attack performance.
With $B_{distill}=16$, only 8bits are required to be flipped to achieve 8.72\% of accuracy 
on ResNet-20 with CIFAR-10 dataset.
It showed similar results on the other benchmarks and DNN models
providing $B_{distill}=16$ to be the best choice for the ZeBRA attack.
Thus, we selected $B_{distill}=16$, $B_{attack}=64$, $\lambda_{CE}=0.2$, $\lambda_{Distill}=0.1$, and $\epsilon_{loss}=10$
when generating $\mathbf{X}_{attack}^d$ for the entire experiments.

\begin{table}[h]
    \centering
    \caption{The attack performance of ZeBRA at various distill batch sizes
    of the distilled target data ($B_{attack}$ = 64, $N_{b}^\text{max}$ = 40)}\vspace{2mm}
    \label{tab:zebra_bsize}
    \scalebox{0.7}{
    \begin{tabular}{cccccc}
    \specialrule{.2em}{.1em}{.1em} 
    \textbf{\begin{tabular}[c]{@{}c@{}}Distill Batch\\ Size ($B_{distill}$)\end{tabular}} & \textbf{\begin{tabular}[c]{@{}c@{}}Bit Flips\\ (Best)\end{tabular}} & \textbf{\begin{tabular}[c]{@{}c@{}} Bit Flips\\ (Mean)\end{tabular}} & \textbf{\begin{tabular}[c]{@{}c@{}}Bit Flips\\ (Worst)\end{tabular}} & \textbf{\begin{tabular}[c]{@{}c@{}}Acc. {[}\%{]}\\ (Best)\end{tabular}} & \textbf{\begin{tabular}[c]{@{}c@{}}Acc. {[}\%{]}\\ (Mean)\end{tabular}} \\ \hline\hline
    \textbf{8}           & 8            & 14.78             & 30                 & 8.77           & 9.93        \\ 
    \textbf{16}          & \textbf{8}   & \textbf{11.56}    & 33                 & 8.72           & 9.58        \\ 
    \textbf{32}          & 17           & 33.40             & 40                 & 9.99           & 10.02       \\ 
    \textbf{64}          & 19           & 35.98             & 40                 & 10.00          & 10.01       \\ 
    \specialrule{.2em}{.1em}{.1em} 
    \end{tabular}}
\end{table}

\section{Visualization of Distilled Target Data}
\Figref{fig:distilled_target_image_imagenet} compares the actual data and the distilled 
target data synthesized by setting \{$\lambda_{CE}=0.2$, $\lambda_{Distill}=0.1$, 
$\epsilon_{loss}=10$, $B_{distill}=16$\} on ResNet-18.
We provide two distilled target data for each image sample trained by the ZeBRA algorithm.
Although synthesized data are visually far from the real data,
they preserve the statistics of batch normalization layers for more efficient attack.
The visualization of the input features to the last layer via
t-SNE~\cite{van2008visualizing} shows that the real and distilled target data of 
the same target label are clustered nearby in the projected feature space
by adding the distilled loss ($\lambda_{Distill}=0.1$; right of Figure~\ref{fig:tsne}) when generating the data.
With $\lambda_{Distill}=0.0$ (left of Figure~\ref{fig:tsne}), the distilled target data on some target labels do not cluster near the real data.
It empirically implies that utilizing only the cross-entropy loss is not 
sufficient to guide the distilled target data to be projected in the desired feature space at the classifier.
The effectiveness of the ZeBRA attack on various $\lambda_{CE}$ and $\lambda_{Distill}$
is reported in Section 4.1.

\begin{figure}[htbp]
    \begin{minipage}[b]{\linewidth}
        \centering
        \includegraphics[width=3in]{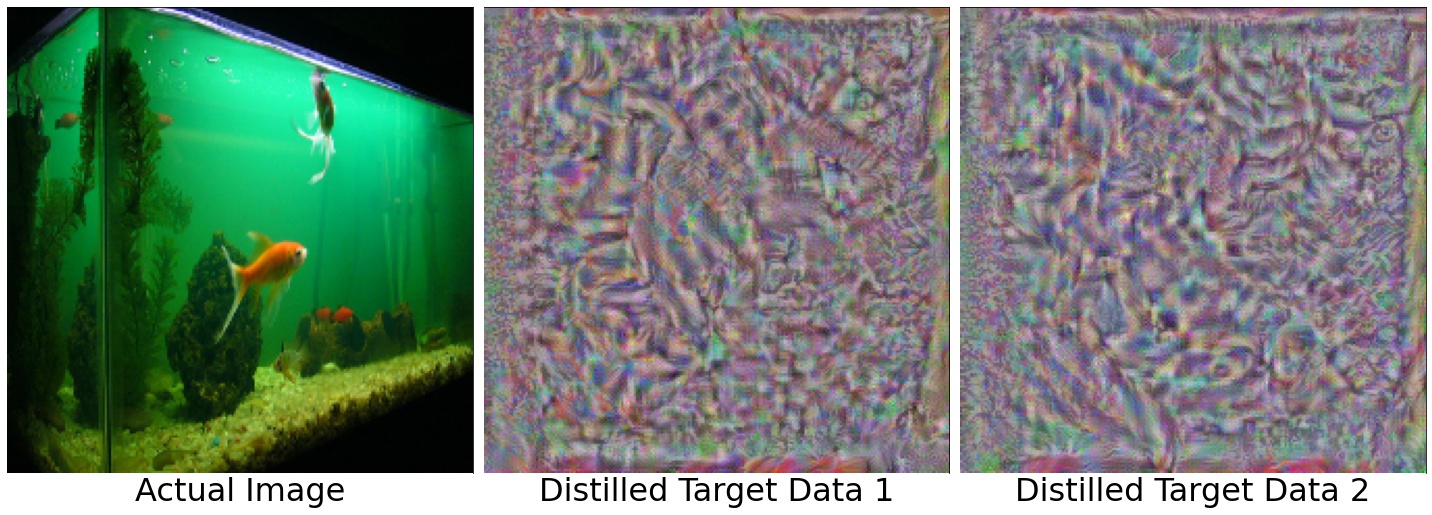}\\
        (a) Gold Fish
    \end{minipage}
    \begin{minipage}[b]{\linewidth}
        \centering
        \includegraphics[width=3in]{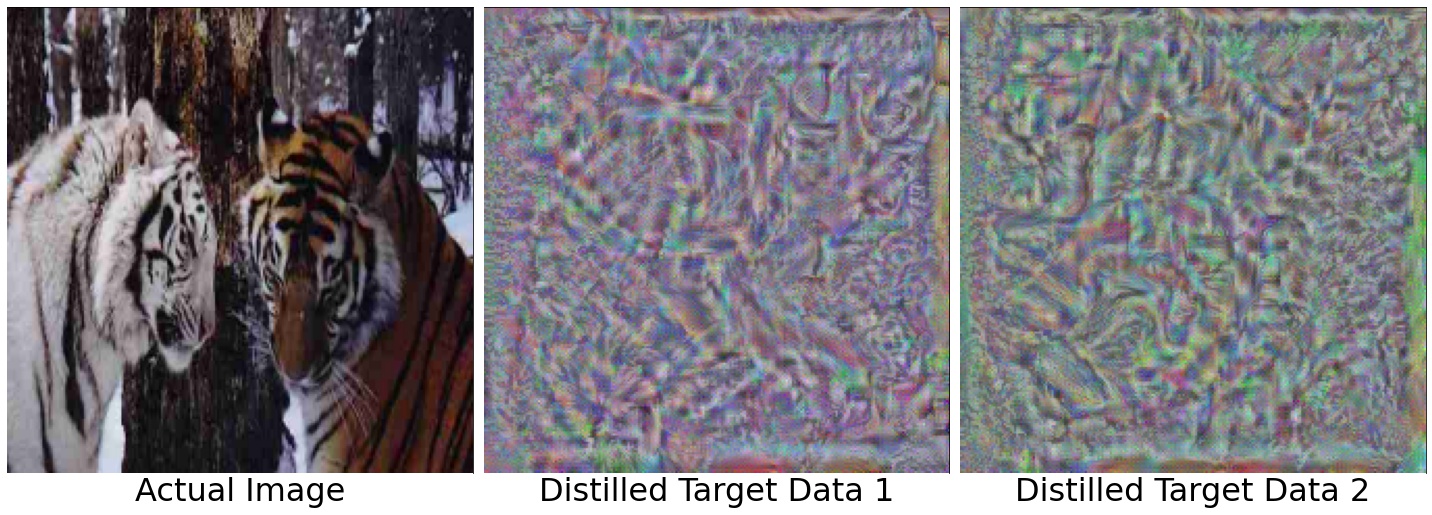}\\
        (b) Tiger
    \end{minipage}
    \vspace{2mm}
    \begin{minipage}[b]{\linewidth}
        \centering
        \includegraphics[width=3in]{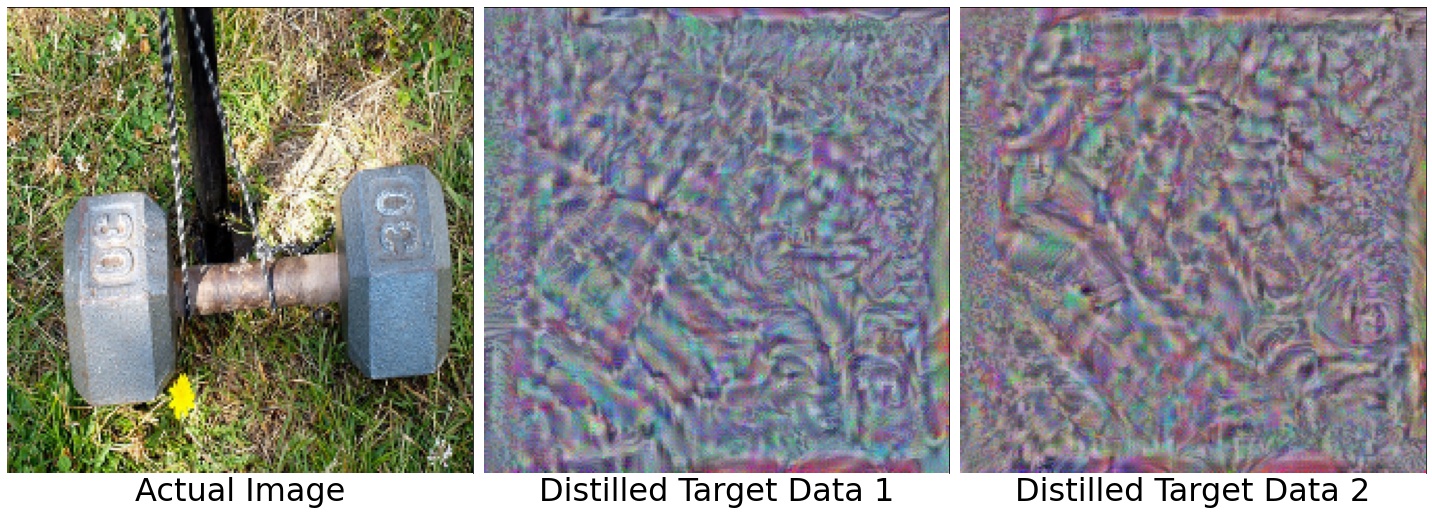}\\
        (c) Dumbbell
    \end{minipage}
    \begin{minipage}[b]{\linewidth}
        \centering
        \includegraphics[width=3in]{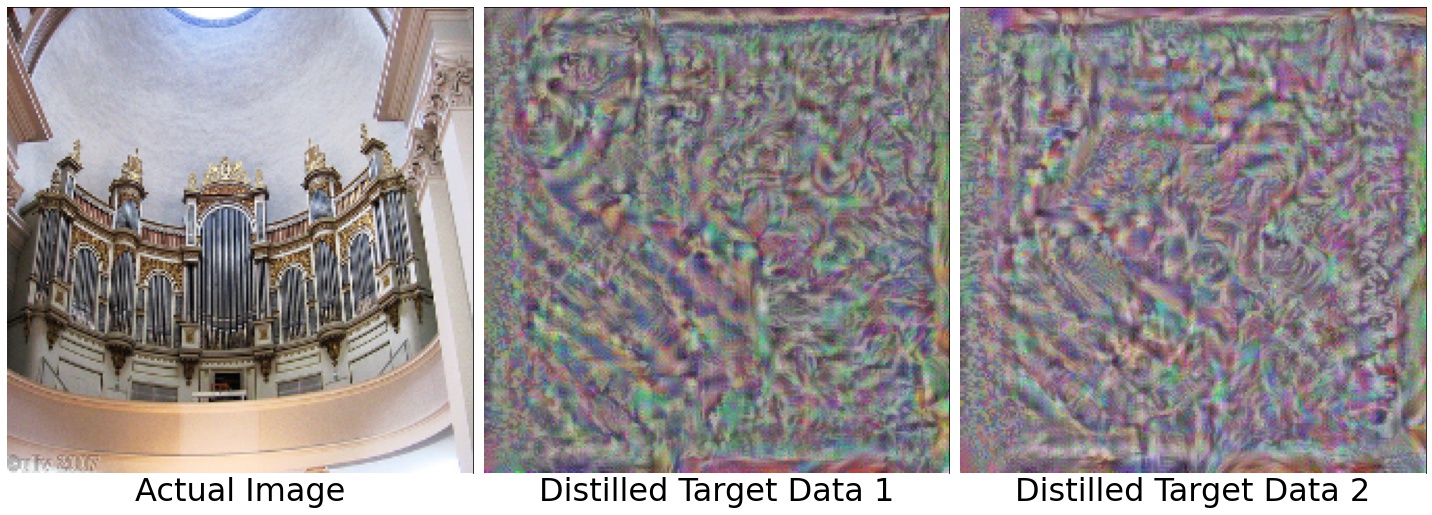}\\
        (d) Organ
    \end{minipage}
    \caption{Comparison between the actual data and distilled target data 
    ($\lambda_{CE}=0.2$, $\lambda_{Distill}=0.1$, 
        $\epsilon_{loss}=10$, $B_{distill}=16$) tested on ResNet-18.}
    \label{fig:distilled_target_image_imagenet}
\end{figure}

\begin{figure}[htbp]
    \centering
    \includegraphics[scale=0.64]{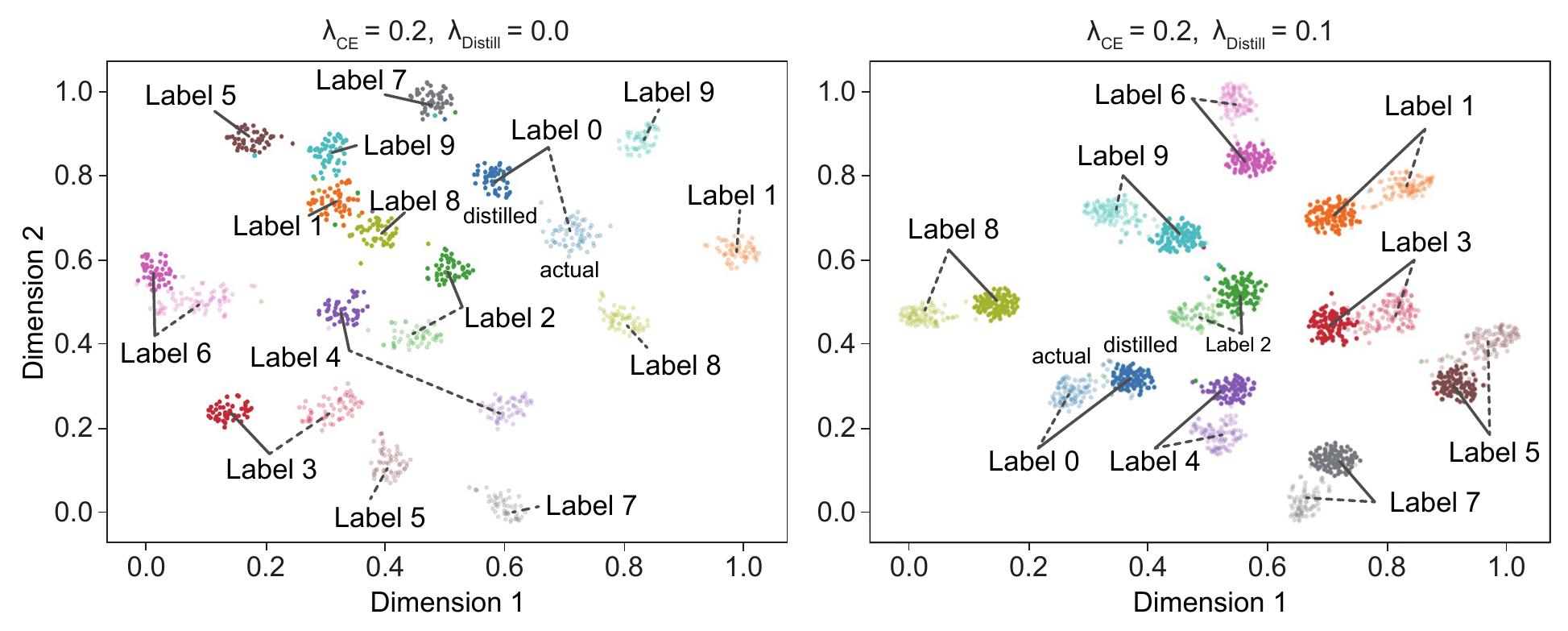}
    \caption{The t-SNE of the actual dataset and the distilled target data
        at the classifier of ResNet-20 on CIFAR-10 with different hyper-parameters:
        (left) $\lambda_{CE}=0.2$, $\lambda_{Distill}=0.0$, $\epsilon_{loss}=10$
        and (right) $\lambda_{CE}=0.2$, $\lambda_{Distill}=0.1$, $\epsilon_{loss}=10$.}
    \label{fig:tsne}
\end{figure}

%
%

\section{Evaluation on Mobile-friendly DNN Models}
As discussed in Section 4.2.2, a mobile-friendly model can be significantly damaged by
the adversarial weight attack due to the existence of pointwise convolutions. 
To support the statement that mobile-friendly DNN models are more vulnerable, 
we evaluated the attack performance of the ZeBRA on MobileNetV2~\cite{mobilenetv2},
ShuffleNetV2~\cite{ma2018shufflenet}, and MnasNet~\cite{tan2019mnasnet}.
As summarized in \Tabref{tab:compare_imagenet_mobile}, the ZeBRA attack on 
most mobile-friendly DNN models results in <0.2\% Top-1 accuracy even by flipping one bit 
(3 bits are required for ShuffleNetV2 1.0).
\Tabref{tab:mobile_distill_val} shows the attack performance of the ZeBRA 
when evaluated by the distilled validation data as discussed in Section 4.3.
\begin{table*}[!]
\centering
\caption{The comparison of the attack performance on mobile-friendly DNN models 
        between the BFA and the proposed ZeBRA on ImageNet dataset}\vspace{2mm}
\label{tab:compare_imagenet_mobile}
\scalebox{0.6}{
\begin{tabular}{ccccccccccc}
\specialrule{.2em}{.1em}{.1em}
\multirow{2}{*}{\textbf{Model}}       & \multirow{2}{*}{\textbf{Accuracy}} & \multirow{2}{*}{\textbf{\begin{tabular}[c]{@{}c@{}}Original\\ Accuracy\\ {[}\%{]}\end{tabular}}} & \multicolumn{4}{c}{\textbf{BFA}}                                                                                                                                                                                                                                                                      & \multicolumn{4}{c}{\textbf{ZeBRA}}                                                                                                                                                                                                                                                                    \\ \cline{4-11} 
                                      &                                    &                                                                                                & \textbf{\begin{tabular}[c]{@{}c@{}}Bit Flips\\ (Best)\end{tabular}} & \textbf{\begin{tabular}[c]{@{}c@{}}Bit Flips\\ (Worst)\end{tabular}} & \textbf{\begin{tabular}[c]{@{}c@{}}Mean\\/ Stdev\end{tabular}}          & \textbf{\begin{tabular}[c]{@{}c@{}}Avg. Accuracy \\ After Attack\end{tabular}} & \textbf{\begin{tabular}[c]{@{}c@{}}Bit Flips\\ (Best)\end{tabular}} & \textbf{\begin{tabular}[c]{@{}c@{}}Bit Flips\\ (Worst)\end{tabular}} & \textbf{\begin{tabular}[c]{@{}c@{}}Mean\\/ Stdev\end{tabular}}          & \textbf{\begin{tabular}[c]{@{}c@{}}Avg. Accuracy \\ After Attack\end{tabular}} \\ \hline\hline

\multirow{2}{*}{\textbf{MobileNetV2}} & \textbf{Top-1}                     & 71.14                                                                                          & \multirow{2}{*}{1}                                                  & \multirow{2}{*}{8}                                                   & \multirow{2}{*}{\begin{tabular}[c]{@{}c@{}}2.65\\/ 0.031\end{tabular}} & 0.14                                                                           & \multirow{2}{*}{1}                                                  & \multirow{2}{*}{2}                                                   & \multirow{2}{*}{\begin{tabular}[c]{@{}c@{}}1.68\\/ 0.014\end{tabular}} & 0.12                                                                           \\
                                      & \textbf{Top-5}                     & 90.01                                                                                          &                                                                     &                                                                      &                                                                         & 0.66                                                                           &                                                                     &                                                                      &                                                                         & 0.61                                                                           \\
                         \hline
\multirow{2}{*}{\textbf{ShuffleNetV2 0.5}} & \textbf{Top-1}                     & 59.36                                                                                          & \multirow{2}{*}{1}                                                  & \multirow{2}{*}{4}                                                   & \multirow{2}{*}{\begin{tabular}[c]{@{}c@{}}2.69\\/ 0.62\end{tabular}} & 0.13                                                                           & \multirow{2}{*}{1}                                                  & \multirow{2}{*}{3}                                                   & \multirow{2}{*}{\begin{tabular}[c]{@{}c@{}}2.60\\/ 0.48\end{tabular}} & 0.12                                                                           \\
                                      & \textbf{Top-5}                     & 81.07                                                                                          &                                                                     &                                                                      &                                                                         & 1.01                                                                           &                                                                     &                                                                      &                                                                         & 0.91                                                                           \\
                         \hline
\multirow{2}{*}{\textbf{ShuffleNetV2 1.0}} & \textbf{Top-1}                     & 68.68                                                                                          & \multirow{2}{*}{2}                                                  & \multirow{2}{*}{5}                                                   & \multirow{2}{*}{\begin{tabular}[c]{@{}c@{}}3.60\\/ 0.80 \end{tabular}} & 0.13                                                                           & \multirow{2}{*}{3}                                                  & \multirow{2}{*}{3}                                                   & \multirow{2}{*}{\begin{tabular}[c]{@{}c@{}}3.00\\/ 0.00\end{tabular}} & 0.16                                                                           \\
                                      & \textbf{Top-5}                     & 87.92                                                                                          &                                                                     &                                                                      &                                                                         & 0.64                                                                           &                                                                     &                                                                      &                                                                         & 0.77                                                                           \\
                         \hline
\multirow{2}{*}{\textbf{MnasNet 0.5}} & \textbf{Top-1}                     & 66.31                                                                                          & \multirow{2}{*}{1}                                                  & \multirow{2}{*}{5}                                                   & \multirow{2}{*}{\begin{tabular}[c]{@{}c@{}}2.25\\/ 0.99 \end{tabular}} & 0.12                                                                           & \multirow{2}{*}{1}                                                  & \multirow{2}{*}{3}                                                   & \multirow{2}{*}{\begin{tabular}[c]{@{}c@{}}1.80\\/ 0.68\end{tabular}} & 0.12                                                                           \\
                                      & \textbf{Top-5}                     & 86.76                                                                                          &                                                                     &                                                                      &                                                                         & 0.96                                                                           &                                                                     &                                                                      &                                                                         & 0.89                                                                           \\
                         \hline
\multirow{2}{*}{\textbf{MnasNet 1.0}} & \textbf{Top-1}                     & 72.34                                                                                          & \multirow{2}{*}{1}                                                  & \multirow{2}{*}{3}                                                   & \multirow{2}{*}{\begin{tabular}[c]{@{}c@{}}1.60\\/ 0.66 \end{tabular}} & 0.12                                                                           & \multirow{2}{*}{1}                                                  & \multirow{2}{*}{2}                                                   & \multirow{2}{*}{\begin{tabular}[c]{@{}c@{}}1.10\\/ 0.30\end{tabular}} & 0.12                                                                           \\
                                      & \textbf{Top-5}                     & 90.73                                                                                          &                                                                     &                                                                      &                                                                         & 0.90                                                                           &                                                                     &                                                                      &                                                                         & 1.02                                                                           \\
                         \specialrule{.2em}{.1em}{.1em}
\end{tabular}}
\end{table*}

\begin{table*}[htbp]
\centering
\caption{The attack performance of the ZeBRA when evaluated by the distilled validation data on mobile-friendly DNN models}\vspace{2mm}
\label{tab:mobile_distill_val}
\scalebox{0.7}{
\begin{tabular}{cccccc}
\specialrule{.2em}{.1em}{.1em}
\multirow{2}{*}{\textbf{Model}} & \multicolumn{5}{c}{\textbf{ZeBRA w/ Distilled Validation Data (ImageNet)}}                                                                                                                                                                                                                                        \\ \cline{2-6}  
                                & \textbf{\begin{tabular}[c]{@{}c@{}}Bit Flips\\ (Best)\end{tabular}} & \textbf{\begin{tabular}[c]{@{}c@{}}Bit Flips\\ (Worst)\end{tabular}} & \textbf{\begin{tabular}[c]{@{}c@{}}Mean\\ / Stdev\end{tabular}} & \textbf{\begin{tabular}[c]{@{}c@{}}Top-1 Accuracy\\ (Min / Avg) \end{tabular}} & \textbf{\begin{tabular}[c]{@{}c@{}}Top-5 Accuracy\\ (Min / Avg) \end{tabular}} \\ \hline\hline
\textbf{MobileNetV2}            & 1                                                                   & 4                                                                    & 2.40 / 0.80                                                       & 0.10 / 1.10      & 0.84 / 5.70                                                             \\
\textbf{ShuffleNetV2 0.5}            & 1                                                                   & 3                                                                    & 1.85 / 0.65                                                       & 0.11 / 1.38      & 0.64 / 3.85                                                             \\ 

\textbf{ShuffleNetV2 1.0}            & 2                                                                   & 3                                                                    & 2.21 / 0.41                                                       & 0.15 / 2.71      & 0.62 / 6.70                                                             \\ 

\textbf{MnasNetV2 0.5}            & 2                                                                   & 3                                                                    & 2.05 / 0.21                                                       & 0.092 / 1.28      & 0.46 / 4.43                                                             \\ 

\textbf{MnasNetV2 1.0}            & 2                                                                   & 3                                                                    & 2.27 / 0.44                                                       & 0.10 / 7.57      & 0.83 / 11.47                                                             \\ 
\specialrule{.2em}{.1em}{.1em}
\end{tabular}}\vspace{-2mm}
\end{table*}

\section{Evaluation of ZeBRA on Additional Datasets}

We conducted additional experiments to show that DNN models trained on medical datasets 
are also vulnerable to ZeBRA attack. 
First, brain tumor dataset~\cite{cheng_2017} consists of 2D MRI slice images and 
there exist three different types of brain tumors to be classified.
We use the same pre-processing on training data as presented in~\cite{swati2019brain}.
The tested DNN models are ResNet-18 and MobileNetV2.
Compared to the BFA attack, the ZeBRA destroys the trained model on the brain tumor dataset
with 7.6\% less number of bit flips on average (Table~\ref{tab:compare_mri}).
Here, the target accuracy is set to 35\% which means that the model has turned into a random predictor.
Another medical dataset under test is a skin disease dataset called DermNet dataset~\cite{dermnet}.
It contains 23 classes and 15,557 training images and 4,002 test images. 
We trained ResNet-50 and MobileNetV2 on the DermNet dataset and performed both the BFA and ZeBRA attacks
for the comparison.
As a result, the ZeBRA successfully destroys DNN models with 22.5\% less number bit flips on average (Table~\ref{tab:compare_skin}).
According to this additional evaluation on two medical datasets,
we can generalize the effectiveness of the ZeBRA attack on a wide range of applications.

\begin{table*}[t]
\centering
\caption{The comparison of the attack performance between the BFA and the proposed ZeBRA on the brain tumor dataset ($B_{attack}=32$, $A_{target}=35\%$)}\vspace{2mm}
\label{tab:compare_mri}
\scalebox{0.64}{
\begin{tabular}{ccccccccccc}
\specialrule{.2em}{.1em}{.1em}
\multirow{2}{*}{\textbf{Model}}       &  \multirow{2}{*}{\textbf{\begin{tabular}[c]{@{}c@{}}Original\\ Accuracy\\ {[}\%{]}\end{tabular}}} & \multicolumn{4}{c}{\textbf{BFA}}                                                                             & \multicolumn{4}{c}{\textbf{ZeBRA}}                                                                                                                                                                                                                                                                    \\ \cline{3-10} 
                                      &                         &                                                                                                          \textbf{\begin{tabular}[c]{@{}c@{}}Bit Flips\\ (Best)\end{tabular}} & \textbf{\begin{tabular}[c]{@{}c@{}}Bit Flips\\ (Worst)\end{tabular}} & \textbf{\begin{tabular}[c]{@{}c@{}}Mean\\/ Stdev\end{tabular}}          & \textbf{\begin{tabular}[c]{@{}c@{}}Avg. Accuracy \\ After Attack\end{tabular}} & \textbf{\begin{tabular}[c]{@{}c@{}}Bit Flips\\ (Best)\end{tabular}} & \textbf{\begin{tabular}[c]{@{}c@{}}Bit Flips\\ (Worst)\end{tabular}} & \textbf{\begin{tabular}[c]{@{}c@{}}Mean\\/ Stdev\end{tabular}}          & \textbf{\begin{tabular}[c]{@{}c@{}}Avg. Accuracy \\ After Attack\end{tabular}} \\ \hline\hline

\multirow{2}{*}{\textbf{ResNet-18}} &  \multirow{2}{*}{98.58}                                                                                          & \multirow{2}{*}{4}                                                  & \multirow{2}{*}{9}                                                   & \multirow{2}{*}{\begin{tabular}[c]{@{}c@{}}6.6\\/ 1.66\end{tabular}} & \multirow{2}{*}{33.85}                                                                           & \multirow{2}{*}{4}                                                  & \multirow{2}{*}{8}                                                   & \multirow{2}{*}{\begin{tabular}[c]{@{}c@{}}6.3\\/ 0.71\end{tabular}} & \multirow{2}{*}{34.02}                                                                           \\ \\
\multirow{2}{*}{\textbf{MobileNetV2}} &  \multirow{2}{*}{97.87}                                                                                          & \multirow{2}{*}{1}                                                  & \multirow{2}{*}{3}                                                   & \multirow{2}{*}{\begin{tabular}[c]{@{}c@{}}1.66\\/ 0.66\end{tabular}} & \multirow{2}{*}{33.53}                                                                           & \multirow{2}{*}{1}                                                  & \multirow{2}{*}{2}                                                   & \multirow{2}{*}{\begin{tabular}[c]{@{}c@{}}1.33\\/ 0.42\end{tabular}} & \multirow{2}{*}{33.59}         
\\ \\
                        \specialrule{.2em}{.1em}{.1em}
\end{tabular}}
\end{table*}

\begin{table*}[t]
\centering
\caption{The comparison of the attack performance between the BFA and the proposed ZeBRA on the skin disease dataset ($B_{attack}=32$, $A_{target}=5\%$)}\vspace{2mm}
\label{tab:compare_skin}
\scalebox{0.64}{
\begin{tabular}{ccccccccccc}
\specialrule{.2em}{.1em}{.1em}
\multirow{2}{*}{\textbf{Model}}       &  \multirow{2}{*}{\textbf{\begin{tabular}[c]{@{}c@{}}Original\\ Accuracy\\ {[}\%{]}\end{tabular}}} & \multicolumn{4}{c}{\textbf{BFA}}                                                                             & \multicolumn{4}{c}{\textbf{ZeBRA}}                                                                                                                                                                                                                                                                    \\ \cline{3-10} 
                                      &                         &                                                                                                          \textbf{\begin{tabular}[c]{@{}c@{}}Bit Flips\\ (Best)\end{tabular}} & \textbf{\begin{tabular}[c]{@{}c@{}}Bit Flips\\ (Worst)\end{tabular}} & \textbf{\begin{tabular}[c]{@{}c@{}}Mean\\/ Stdev\end{tabular}}          & \textbf{\begin{tabular}[c]{@{}c@{}}Avg. Accuracy \\ After Attack\end{tabular}} & \textbf{\begin{tabular}[c]{@{}c@{}}Bit Flips\\ (Best)\end{tabular}} & \textbf{\begin{tabular}[c]{@{}c@{}}Bit Flips\\ (Worst)\end{tabular}} & \textbf{\begin{tabular}[c]{@{}c@{}}Mean\\/ Stdev\end{tabular}}          & \textbf{\begin{tabular}[c]{@{}c@{}}Avg. Accuracy \\ After Attack\end{tabular}} \\ \hline\hline

\multirow{2}{*}{\textbf{ResNet-50}} &  \multirow{2}{*}{63.23}                                                                                          & \multirow{2}{*}{3}                                                  & \multirow{2}{*}{29}                                                   & \multirow{2}{*}{\begin{tabular}[c]{@{}c@{}}9.7\\/ 6.61\end{tabular}} & \multirow{2}{*}{4.68}                                                                           & \multirow{2}{*}{5}                                                  & \multirow{2}{*}{10}                                                   & \multirow{2}{*}{\begin{tabular}[c]{@{}c@{}}7.53\\/ 1.09\end{tabular}} & \multirow{2}{*}{4.71}                                                                           \\ \\
\multirow{2}{*}{\textbf{MobileNetV2}} &  \multirow{2}{*}{54.05}                                                                                          & \multirow{2}{*}{2}                                                  & \multirow{2}{*}{9}                                                   & \multirow{2}{*}{\begin{tabular}[c]{@{}c@{}}3.5\\/ 2.04\end{tabular}} & \multirow{2}{*}{4.58}                                                                           & \multirow{2}{*}{2}                                                  & \multirow{2}{*}{4}                                                   & \multirow{2}{*}{\begin{tabular}[c]{@{}c@{}}2.7\\/ 0.64\end{tabular}} & \multirow{2}{*}{4.82}         
\\ \\
                        \specialrule{.2em}{.1em}{.1em}
\end{tabular}}
\end{table*}

\section{ZeBRA on Fortified DNN Models}

In~\cite{harness}, authors found that the BFA tends to
select and flip MSBs making small weights large values (e.g., from 0 to $\pm$128).
The same observations were made when we perform the ZeBRA on all DNNs.
Based on this observation, piece-wise clustering of weight parameters has been proposed
to avoid the large shift in weight values~\cite{harness}.
To perform the piece-wise clustering,
the following penalty term is added to CE loss during the training:
\begin{equation}\label{eq:pc_penalty}
\lambda_{cluster}\cdot \sum_{l=0}^{L-1}(||\theta_l^+ - \mathbb{E}(\theta_l^+)||_2
+||\theta_l^- - \mathbb{E}(\theta_l^-)||_2),
\end{equation}
where $\lambda_{cluster}$ is the cluster coefficient and
$\theta_{l}^+$ (or $\theta_{l}^-$) denotes the positive (or negative) weight parameters
at $l$-th layer.
By adding~\eqnref{eq:pc_penalty}, the weight distribution of a DNN model changes
from Gaussian to bimodal distribution.

We call the DNN model with piece-wise clustering `the fortified DNN model'.
After training the fortified ResNet-20 on CIFAR-10,
we compared the attack performance between the BFA and ZeBRA
at various quantization levels ($N_Q$) and cluster coefficients ($\lambda_{cluster}$).
The Top-1 accuracy prior to the adversarial weight attack
is reported in \Tabref{tab:defense}.
As shown in \Figref{fig:defense},
the average number of bit flips to reach 10\% Top-1 accuracy
using the ZeBRA (22.4 bits) is 1.7$\times$ lower than the BFA (38.7 bits).
Thus, the ZeBRA shows better attack performance even at the fortified DNNs.

\begin{table}[h]
\centering
\caption{Top-1 test accuracy of fortified ResNet-20 on CIFAR-10}\vspace{2mm}
\label{tab:defense}
\scalebox{0.76}{
\begin{tabular}{ccccc}
\specialrule{.2em}{.1em}{.1em}
\textbf{$N_Q$} & \textbf{$\lambda_{cluster}$ = 0} & \textbf{0.0001} & \textbf{0.0005} & \textbf{0.001} \\ \hline\hline
\textbf{8bit}  & 92.41          & 92.06              & 91.64               & 91.06               \\
\textbf{6bit}  & 92.18          & 91.78              & 91.29               & 90.87               \\
\textbf{4bit}  & 87.59          & 88.87              & 90.23               & 89.33               \\ \specialrule{.2em}{.1em}{.1em}
\end{tabular}}
\end{table}

\begin{figure}[tbhp]
    \centering
    \includegraphics[scale=0.7]{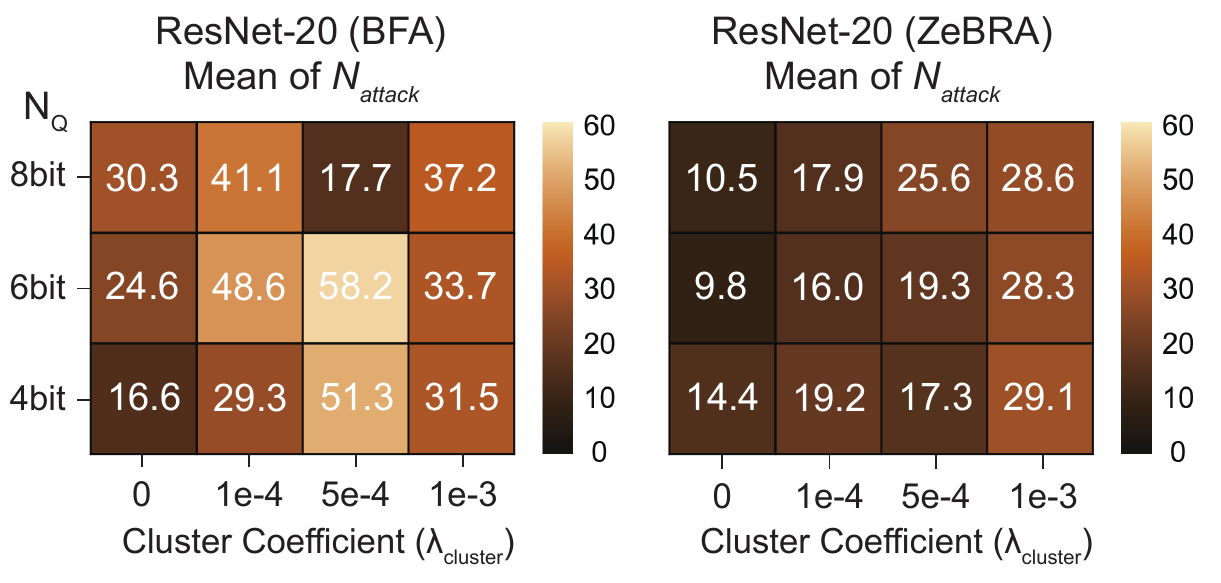}
    \caption{The average number of bit flips to destroy the fortified ResNet-20 on CIFAR-10
        using the BFA or ZeBRA.}
    \label{fig:defense}\vspace{-1mm}
\end{figure}

\bibliography{egbib}

\end{alphasection}